\documentclass[conference]{IEEEtran}
\IEEEoverridecommandlockouts
\usepackage{cite}
\usepackage{amsmath,amssymb,amsfonts,amsthm}
\usepackage{graphicx}
\usepackage{textcomp}
\usepackage{xcolor}

\usepackage{algorithm}
\usepackage{array}
\usepackage{textcomp}
\usepackage{url}
\usepackage{verbatim}
\usepackage[algo2e]{algorithm2e}
\usepackage[noend]{algorithmic}

\newtheorem{theorem}{Theorem}

\usepackage{diagbox}
\usepackage[export]{adjustbox}

\usepackage{makecell}
\usepackage{multirow}
\usepackage{soul} 
\usepackage{graphics}
\usepackage{graphicx}

\usepackage{subfigure}
\usepackage{subfig}

\def\BibTeX{{\rm B\kern-.05em{\sc i\kern-.025em b}\kern-.08em
    T\kern-.1667em\lower.7ex\hbox{E}\kern-.125emX}}
\begin{document}

\title{A Client-level Assessment of  Collaborative Backdoor Poisoning in Non-IID Federated Learning
}



\author{\IEEEauthorblockN{*Phung Lai}
\IEEEauthorblockA{
\textit{University at Albany}\\
Albany, New York, USA \\
lai@albany.edu}
\and
\IEEEauthorblockN{*Guanxiong Liu}
\IEEEauthorblockA{
\textit{Meta Inc.}\\
Menlo Park, California, USA \\
gl236@njit.edu}
\and
\IEEEauthorblockN{ NhatHai Phan }
\IEEEauthorblockA{
\textit{New Jersey Institute of Technology}\\
Newark, New Jersey, USA \\
phan@njit.edu }
\and
\IEEEauthorblockN{Issa Khalil}
\IEEEauthorblockA{
\textit{Qatar Computing Research Institute}\\
Doha, Qatar \\
ikhalil@hbku.edu.qa}
\and
\IEEEauthorblockN{Abdallah Khreishah}
\IEEEauthorblockA{\textit{New Jersey Institute of Technology} \\
Newark, New Jersey, USA \\
abdallah@njit.edu}
\and
\IEEEauthorblockN{Xintao Wu}
\IEEEauthorblockA{\textit{University of Arkansas}\\
Fayetteville, Arkansas, USA \\
xintaowu@uark.edu}
}

\maketitle
\begingroup\renewcommand\thefootnote{\textsection}
\footnotetext{* Equal contribution}
\endgroup

\begin{abstract}
 Federated learning (FL) enables collaborative model training using decentralized private data from multiple clients. While FL has shown robustness against poisoning attacks with basic defenses, our research reveals new vulnerabilities stemming from non-independent and identically distributed (non-IID) data among clients. These vulnerabilities pose a substantial risk of model poisoning in real-world FL scenarios.

To demonstrate such vulnerabilities, we develop a novel collaborative backdoor poisoning attack called \textsc{CollaPois}. In this attack, we distribute a single pre-trained model infected with a Trojan to a group of compromised clients. These clients then work together to produce malicious gradients, causing the FL model to consistently converge towards a low-loss region centered around the Trojan-infected model. Consequently, the impact of the Trojan is amplified, especially when the benign clients have diverse local data distributions and scattered local gradients. 
\textsc{CollaPois} stands out by achieving 
its goals while involving only a limited number of compromised clients, setting it apart from existing attacks. Also, \textsc{CollaPois} effectively avoids noticeable shifts or degradation in the FL model's performance on legitimate data samples, allowing it to operate stealthily and evade detection by advanced
robust FL algorithms.

Thorough theoretical analysis and experiments conducted on various benchmark datasets demonstrate the superiority of \textsc{CollaPois} compared to state-of-the-art backdoor attacks. Notably, \textsc{CollaPois} bypasses existing backdoor defenses, especially in scenarios where clients possess diverse data distributions. Moreover, the results show that \textsc{CollaPois} remains effective even when involving a small number of compromised clients. Notably, clients whose local data is closely aligned with compromised clients experience higher risks of backdoor infections.
\end{abstract}


\section{Introduction}

Recent and emerging data privacy regulations \cite{europe2018reg} challenge machine learning (ML) applications that collect sensitive user data on centralized servers. Federated learning (FL) \cite{mcmahan2017communication} addresses this by enabling collaborative model training without sharing raw data. However, FL often faces performance disparities due to diverse data distributions across clients \cite{zhu2021federated}.

 
To address this challenge, personalization in FL has gained considerable attention. Personalization allows each client's model to adjust to its own unique data distribution, and various methods have been proposed to achieve this
\cite{shamsian2021personalized,fallah2020personalized}. 
In contrast to conventional FL methods, personalized FL methods are more suitable for individual clients, especially in situations where there are significant variations in data distributions across clients. In real-world scenarios, clients often employ FL at different geographical locations or serve different user cohorts \cite{jiang2022flsys,jiang2023zone}. This often results in significant variations in data distribution, making personalized FL a promising choice.

FL has been extensively studied to identify potential adversarial and Trojan vulnerabilities \cite{xie2020dba, bagdasaryan2020backdoor, sun2019can, zhang2022neurotoxin}, given its significance and widespread usage. Despite this, recent research (drawing on various sources) \cite{drawingboard2022} surprisingly shows that FL remains relatively resilient to previously reported poisoning attacks, even when utilizing low-cost robust training algorithms. However, commonly employed aggregate metrics in previous research, such as the average accuracy across all clients, do not sufficiently reflect the individualized impact of proposed attacks or defenses on each client.
This is concerning because a high average accuracy may mask unacceptably low accuracy levels for certain clients, particularly if there is substantial variation in their individual accuracy, which may lead to an underestimation of FL vulnerabilities. 
Ensuring a well-balanced client-level accuracy is of utmost importance as clients actively participate in FL with the expectation of achieving good performance. Consequently, an attack that substantially impacts the performance of a small subset of clients can pose a significant threat to the entire FL system. 
The high diversity in data distribution among clients, which is a key feature of FL, becomes even more pronounced when coupled with personalization techniques. As a result, it becomes imperative to thoroughly assess the effects of attacks and defenses on individual clients.
Concretely, we seek to answer the  question: \textbf{\textit{How many clients are impacted, which ones, and to what extent in terms of attack success rates, and what are the underlying reasons for these impacts?}}

\textbf{Key Contributions.} To close these gaps, we introduce a novel Trojan attack called \textsc{CollaPois}, with the goal of systematically elucidating the connection between the risk of backdoor poisoning and the degree of diversity present in the local data distributions of clients in the context of FL.


In contrast to existing backdoor  attacks, \textsc{CollaPois} leverages the diverse local data distributions of benign clients and the resulting scattered gradients to steer the federated training model towards a pre-trained Trojaned model $X$. As a result, backdoors can be transferred to benign clients' local models  through the Trojan-infected federated training model.

To achieve this, we adopt a coordinated approach involving a group of compromised clients. With this approach, we generate well-aligned malicious gradients, in stark contrast to the scattered gradients contributed by the benign clients. This compels the federated training model to converge within a tightly confined region surrounding the Trojaned model $X$, effectively poisoning the federated training model.

\vspace{-1pt}

Our extensive theoretical and empirical analysis show that \textsc{CollaPois} can significantly lower and bound the number of compromised clients required to successfully carry out backdoor poisoning as a function of the attack stealthiness and the degree of diversity in benign clients' local data distribution. We refer to \textit{\textbf{stealthiness}} as the ability to prevent the server from approximating the Trojaned model $X$ and identifying compromised clients. In fact, \textit{\textbf{the greater the diversity}} in benign clients' local data, \textit{\textbf{the fewer compromised clients are required}} and \textit{\textbf{the more stealthy the successful poisoning}} becomes, and vice-versa. This establishes a new theoretical connection among these three key components in FL. The novelty of \textsc{CollaPois} also stems from its \textbf{\textit{Simplicity}}, as it involves only minor adjustments to classical  poisoning techniques in FL \cite{suciu2018does, li2016data}. These modifications do not necessitate any additional knowledge in the federated training process. Moreover, its \textbf{\textit{Efficiency}} makes it cost-effective for compromised clients to compute malicious local gradients during a training round. Consequently, \textsc{CollaPois} becomes a practical method to expose backdoor risks in FL.
 
Through comprehensive assessments with client-level metrics, we show that \textsc{CollaPois} outperforms existing attacks and adeptly evades various robust federated training and defense algorithms, with and without personalization. Even with only a modest 0.5\% of compromised clients, \textsc{CollaPois} can effectively create a backdoor for 15\% of benign clients with an impressive success rate exceeding 70\% under robust federated training on benchmark datasets.


\section{Background and Related Work}

\subsection{Federated Learning and Personalization}

\textbf{ Federated Learning (FL).} FL is a multi-round protocol between a server and $N$ devices. At each round $t$, the server sends the global model $\theta^t$ to a random subset $M^t$, which trains local models $\theta^t_u$ on data $D_u$ and returns them. The server aggregates these updates using a function $\mathcal{G}$ to produce $\theta^{t+1} = \mathcal{G}({\theta_u^{t+1}}_{u\in M^t})$. FedAvg is a commonly used aggregation method \cite{kairouz2019advances}.

\textbf{Non-IID Data in FL.} One of the fundamental challenges that could impair performance and may introduce vulnerabilities in FL models is the non-independent and identically distributed (non-IID) clients' local data \cite{collins2021exploiting,kairouz2019advances}. 
Non-IID data is characterized by notable distinctions in the local data distributions of clients. In this study, we investigate label distribution skew as one prevalent form of non-IID data \cite{kairouz2019advances}.
Similar to earlier research \cite{luo2021ensemble}, we assume that the distribution of the number of data samples per class (label) within each client adheres to a symmetric Dirichlet distribution denoted as $Dir(\alpha)$, in which the concentration parameter $\alpha$ controls the degree of non-IIDness.
Values of $\alpha$ above 1 favor dense and evenly distributed classes, while values below 1 favor diverse and sparse distributions with data concentrated in fewer classes. Throughout the remainder of this study, ``diversity'' refers to the non-IID level of clients' local data distribution.


\textbf{Personalized FL.}  
 Personalized Federated Learning (PFL) approaches address non-IID challenges by adapting performance to individual clients' local data distributions \cite{shamsian2021personalized, gao2022feddc}. PFL approaches can be broadly categorized into four main research directions: \textbf{(1) Regularization-based} approaches apply penalties to local training to address data distribution drift and reduce discrepancies between local and global model weights \cite{reddi2020adaptive, gao2022feddc}, \textbf{(2) Knowledge distillation} allows the server to combine clients' knowledge via a generator or consensus, enabling clients to use this knowledge as an inductive bias or to train local models using public and private data \cite{li2019fedmd,shamsian2021personalized}, \textbf{(3) Clustering-based} frameworks  assign clients to clusters and aggregate local models within each cluster \cite{ghosh2020efficient,10.1145/3576915.3623212}, and \textbf{(4) Meta-Learning} leverages the concept of meta-training and meta-testing. In meta-training, a flexible initial model is learned using methods like Model Agnostic Meta-Learning, enabling quick adaptation to various tasks. This model maps to the global model, which is then fine-tuned for specific tasks during meta-testing on the client's side.

In this study, we focus on regularization and knowledge distillation-based FL approaches, as they generalize to clustering and meta-learning methods. These methods allow FL models learned for client clusters to adapt to various tasks.

\subsection{Backdoor Attacks}
\label{Backdoor Attacks} 

This section reviews the threat model of FL poisoning attacks. We first introduce \textbf{\textit{attack knowledge}} and \textbf{\textit{attack capability}} to categorize the attacks discussed.


\textbf{Attack Knowledge.}  
Attack knowledge refers to an attacker's awareness of other clients' information. \textbf{\textit{White-box}} knowledge means knowing other clients' updates, while \textbf{\textit{black-box}} one implies no such access, making it more practical.


\textbf{Attack Capability.}  
We classify attack capability into \textbf{\textit{partial capability}}, where the attacker can only inject poisoned data into a subset of clients' training sets, and \textbf{\textit{full capability}}, where the attacker  can control the entire subset of clients (referred to as \textbf{\textit{compromised clients}}  and  manipulate their training process at will $\mathcal{C}$). The attacker with full capability can control compromised clients to coordinate poisoning attacks. Both attackers are practical in FL  \cite{drawingboard2022}.


\textbf{Backdoor Attacks.}
This work focuses on backdoor attacks that misclassify specific inputs while maintaining model accuracy on legitimate data. 
Trojans have emerged as a prominent method for conducting backdoor attacks, as highlighted in previous studies \cite{gu2017badnets,liu2017trojaning}. Trojans involve carefully embedding a specific pattern, such as a brand logo or additional pixels, into legitimate data samples to induce the desired misclassification. Recently, an image warping-based Trojan has been developed, which subtly distorts an image using geometric transformations \cite{nguyen2021wanet}. 
This technique   makes modifications imperceptible to human observers. Importantly, warping techniques enable Trojans to evade commonly used detection methods like Neural Cleanse \cite{wang2019neural}, Fine-Pruning \cite{liu2018fine}, and STRIP \cite{gao2019strip}. The attacker activates the backdoor during the inference phase by applying this Trojan trigger to legitimate data samples. In this study, we employ the warping-based Trojan technique \cite{nguyen2021wanet}.



In FL, backdoor attacks involve compromised clients  controlled by an attacker  to create malicious gradients before sending them to the server. The attacker can apply data poisoning (\textsc{DPois}) and model replacement (\textsc{MRepl}) approaches to generate malicious local gradients. 
In \textsc{DPois} \cite{suciu2018does,li2016data}, compromised clients train on Trojaned datasets to generate malicious local gradients, causing the aggregated model to exhibit backdoor effects.
In \textsc{MRepl} \cite{bagdasaryan2020backdoor}, adversaries create malicious local gradients to partially or entirely replace  the server's aggregated model with a Trojaned model, even after a single training round \cite{fang2020local}. However, \textsc{MRepl} often causes noticeable performance shifts \cite{bagdasaryan2020backdoor}, making detection easier by monitoring abrupt changes across training rounds.


\begin{figure}[t] \vspace{-10pt}
\centering
\subfigure[$0.1\%$ compromised clients]{\includegraphics[scale=0.24]{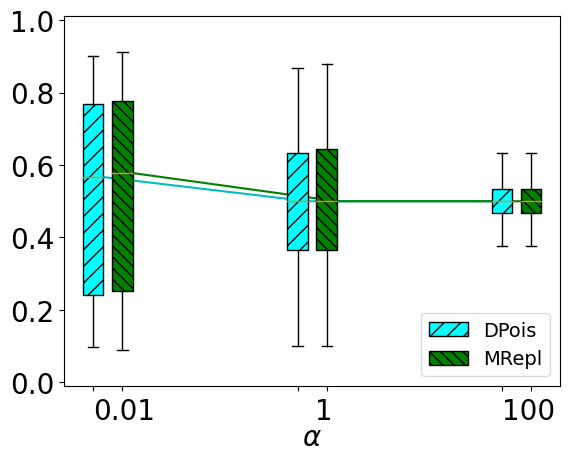}}\hfill
\subfigure[$1\%$ compromised clients]{\includegraphics[scale=0.24]{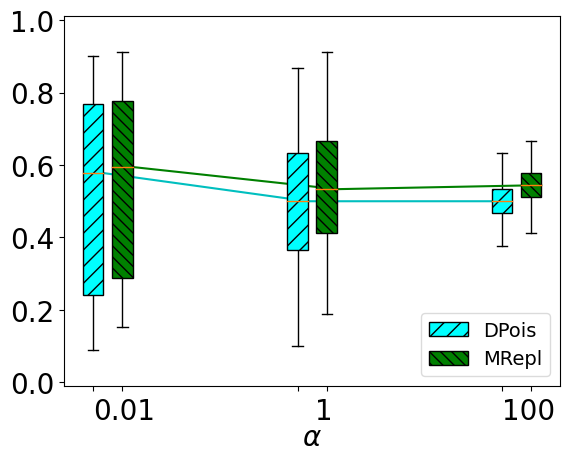}}\hfill \vspace{-5pt}
\caption{\textsc{DPois} and \textsc{MRepl} attacks show modest changes, with $0.1\%$ and $1\%$ compromised clients across  distribution levels.} 
\label{notintuitive}  
\end{figure} 
\setlength{\textfloatsep}{2pt}

\textsc{DPois} and \textsc{MRepl} were not designed to exploit non-IID data for more effective poisoning in FL. Current  methods lack client-level risk insights, threatening FL integrity.
\vspace{-10pt}




\subsection{Defenses against Backdoor Attacks in FL} \label{sec:background-defenses}
Current defense mechanisms against backdoor poisoning attacks in FL can be classified into two categories: \textbf{(1)} Detection of Trojans during inference \cite{wang2019neural,gao2019strip,ma2022beatrix,cheng2023beagle} by identifying or decomposing poisoned samples to clean inputs and triggers. These approaches typically demand computing-intensive resources, given their high computational complexity. This expensive computation overhead hinders their applicability to clients with limited resources in federated learning;
\textbf{(2)} Poisoned update detection \cite{10.1145/3576915.3623212} by checking difference between malicious and benign gradients  using statistical tests; 
and \textbf{(3)} Resilient gradient aggregation to reduce the impact of malicious local gradients during the federated training process \cite{yin2018byzantine,hong2020effectiveness,ozdayi2020defending}. Unlike the first category, the other approaches are efficient and do not incur extra computational costs for clients in FL. Therefore, when discussing defense mechanisms against our attack, we refer to methods that can ensure a robust aggregation process in FL and PFL, effectively preventing backdoor Trojans from being transferred to benign clients.

\textbf{Robust Federated Training.}  Table \ref{tab:compare defenses} (Supplementary\footnote{https://anonymous.4open.science/r/CollaPois/}) summarizes various robust federated training approaches. 
They include coordinate-wise median, geometric median, $\alpha$-trimmed mean, as well as their variants and combinations, as outlined in the literature \cite{yin2018byzantine}. Recently proposed approaches include weight-clipping and noise
addition with certified bounds, ensemble models, differential privacy (DP) optimizers, and adaptive and robust learning rates (RLR) across clients and at the server \cite{hong2020effectiveness,ozdayi2020defending}. Despite differences, existing robust aggregation focuses on analyzing and manipulating the local gradients $\bigtriangleup \theta^{t}_i$, which share the global aggregated model $\theta_t$ as the same root, i.e., $\forall i \in [N]: \bigtriangleup \theta_i^t = \theta_i^t - \theta^t$. The fundamental assumption in these approaches is that the local gradients from compromised clients $\{\bigtriangleup {\theta}^{t}_c\}_{c \in \mathcal{C}}$ and from legitimate clients $\{\bigtriangleup \theta^{t}_i\}_{i \in N \setminus \mathcal{C}}$ are different in terms of magnitude and direction.

\section{Numbers of Compromised Clients and Non-IID}

Our first effort to draw the correlation between the number of compromised clients and non-iid data in backdoor poisoning attacks is launching \textsc{DPois} and \textsc{MRepl} attacks in the Sentiment dataset \cite{go2009twitter}, which has 5,600 clients and 1 million samples.  
Data distribution across clients follows a symmetric Dirichlet distribution with concentration parameter $\alpha \in [0.01, 100]$. We conduct experiments with 0.1\% and 1\% compromised clients under varying non-IID levels, i.e., $\alpha \in [0.01, 100]$. 
As shown in  Fig.~\ref{notintuitive}, there are modest changes observed between $0.1\%$ and $1\%$ compromised clients across different levels of data distribution, i.e., $\alpha \in [0.01, 100]$, in existing \textsc{DPois} and \textsc{MRepl} attacks. This result highlights a gap in understanding the correlation between non-IID data distribution, attack stealthiness, and attack effectiveness, motivating our systematic study of backdoor risks in FL under non-IID data from theoretical and empirical angles.


\section{Collaborative Poisoining Attacks}\label{sec:proposed-attack}



\subsection{Threat Model} 

\begin{figure}[t]
      \centering \vspace{-15pt}
       \includegraphics[scale=0.022]{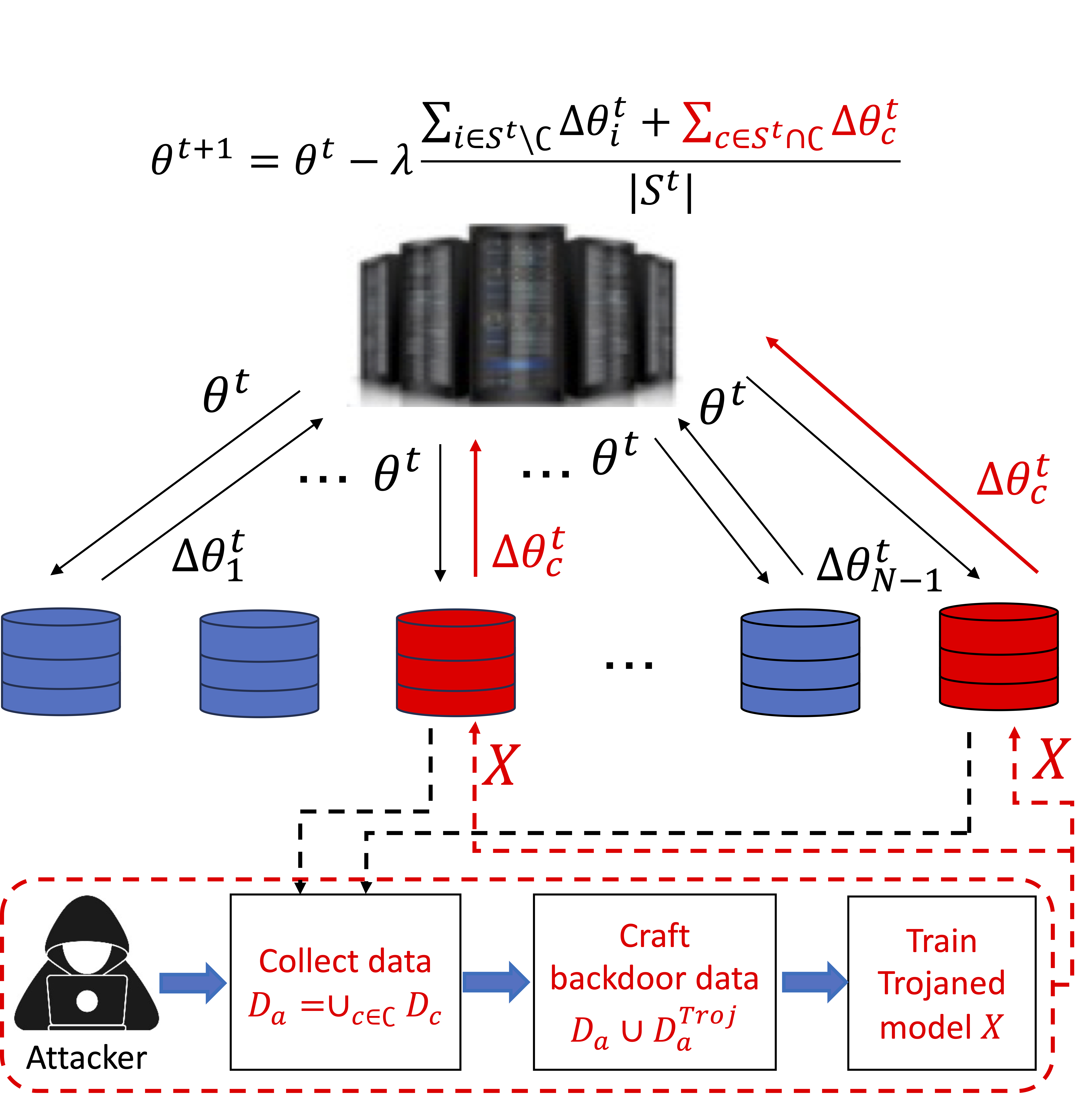} 
      \caption{CollaPois framework. In each training round, compromised clients send malicious gradients (red-solid) to steer the FL model $\theta$ toward a Trojaned model $X$ sent by the attacker (red-dashed). Dashed and solid vectors indicate one-time and multiple training rounds, respectively.}
      \vspace{-5pt}
      \label{ThreatModels}
\end{figure}

We consider black-box poisoning carried out by an attacker with full capability in FL as in Section \ref{Backdoor Attacks}. Fig.~\ref{ThreatModels} shows our threat model. The server is honest and strictly follows the federated training protocol and there is no collusion between the server and the attacker. 
This threat model is crucial for servers and service providers aiming to maximize client model utility, as poisoning can degrade performance, reducing incentives to harm service quality.

The attacker fully controls a set of compromised clients $\mathcal{C}$ participating in the federated training.
The literature considers this attacker as a practical threat when the complement set size, $|\mathcal{C}|$, is small, e.g., 0.01-10\% of clients \cite{drawingboard2022}. The attacker has access to an auxiliary dataset, referred to as $D_a$, which is composed of local datasets collected by the compromised clients, denoted as $D_a = \cup_{c \in \mathcal{C}} D_c$. $D_a$ shares the same downstream task with benign clients. 

The attacker's objective is to manipulate the federated training process by transmitting malicious local gradients through compromised clients to the server, producing backdoored local models that deviate from clean local models in benign clients. 
An optimal backdoored model behaves identically to a clean model for legitimate inputs but provides a prediction of the attacker's choosing when the input contains a backdoor trigger, such as a Trojan \cite{nguyen2021wanet}. 
The attack is more effective as more benign clients are impacted by backdoors, increasing success rates without reducing model utility on legitimate inputs. It also becomes stealthier and more practical when fewer compromised clients are needed, making detection harder.


\setlength{\textfloatsep}{6pt}

\subsection{Collaborative Poisoning (\textsc{CollaPois})}
 
Existing backdoor  attacks have not been designed to exploit scattered gradients that arise due to the non-IID characteristics of clients' data in FL. In contrast to existing attacks, \textsc{CollaPois} investigates model integrity risks caused by scattered gradients and demonstrates how these scattered gradients can be harnessed for potent attack strategies.

The pseudo-code of \textsc{CollaPois} is in Algorithm \ref{Model Transferring Attacks}. First, the attacker poisons the auxiliary data $D_a$ by embedding a backdoor trigger (Trojan) into the data samples and changing their labels to match the attacker's desired prediction. This manipulation results in a collection of perturbed data samples, denoted as $D_a^{Troj}$, which the attacker employs to train a Trojaned model $X$ using a centralized approach (Line 3). The training process minimizes the following objective loss: 

\vspace{-10pt}
\begin{equation}
    X = \arg\min_{\theta_a} L(\theta_a, D_a \cup D_a^{Troj}),  
    \vspace{-5pt}
\end{equation}
where $\theta_a$ is the attacker's model used to train $X$, sharing the same structure as the global model $\theta$, as the attacker learns its structure through the compromised clients.

In this study, the attacker uses WaNet \cite{nguyen2021wanet}, which is one of the state-of-the-art backdoor attacks, for generating the poisoned image data $D_a^{Troj}$. 
WaNet uses image warping-based triggers to create natural and unnoticeable backdoor modifications. Following \cite{nguyen2021wanet}, backdoor images are generated to train the Trojaned model $X$ (samples in Fig.~\ref{cifar10-vis}, Supplementary). For the text data used in the evaluation, we follow existing Trojan attack \cite{alsharadgah2021adaptive}, which use a fixed term as the trigger.

Employing the Trojaned model $X$ shared by the attacker, the compromised clients $\mathcal{C}$ compute their 
malicious local gradients as $\{\bigtriangleup \theta_c^t = X - \theta^t\}_{c \in \mathcal{C}}$ in each training round $t$ (Lines 12 and 13). If a compromised client $c$ is selected with probability $q$ in training round $t$, it sends malicious local gradients $\bigtriangleup \theta_c^t$ to the server upon receiving the latest model update $\theta^t$. The global model is then aggregated and updated as follows: 

\vspace{-5pt}
\begin{equation}
\theta^{t+1} = \theta^t - \lambda \Big( \sum_{i \in S^t \setminus \mathcal{C}} \bigtriangleup \theta^{t}_i + \sum_{c \in S^t \cap \mathcal{C}} \bigtriangleup \theta^{t}_c\Big) /|S_t|. 
\label{PoisFedAvg}
\end{equation}
\vspace{-5pt}


The Trojaned surrogate loss minimized by the attacker and the compromised clients is as follows:
\begin{equation} 
\frac{1}{2} (\sum_{c \in \mathcal{C}} \| X - \theta^* \|_2^2 + \sum_{i \in N \setminus \mathcal{C}} \| \theta^{*}_i - \theta^* \|_2^2),
\label{Trojaned surrogate loss}
\end{equation}
where $\theta^*$ represents the optimal global FL model, and $i \in N \setminus \mathcal{C}$ are the benign clients and their associated loss functions on their legitimate local datasets $D_i$: $\theta^*_{i} =  \arg\min_{\theta_i} L(\theta_i, D_i)$.
In practice, $\theta^{*}_i$ serves as a personalized model for client $i$.

\begin{algorithm}[t]
\small
\caption{Collaborative Poisoning Attack (\textsc{CollaPois})}\label{Model Transferring Attacks}
\begin{algorithmic}[1]
\STATE \textbf{Input}: Number of compromised clients $|\mathcal{C}|$, number of benign clients $|N|-|\mathcal{C}|$, client sampling probability $q$, number of rounds $T$, number of local rounds $K$, server's learning rates $\lambda$, clients' learning rate $\gamma$, a random and dynamic learning rate $\psi \sim \mathcal{U}[a, b]$, and $L_i(B)$ is the loss function $L_i(\theta)$ on a mini-batch $B$
\STATE \textbf{Output}: $\theta$
\STATE Attacker trains a Trojaned model $X= \arg\min_{\theta_a} L(\theta_a, D_a \cup D_a^{Troj})$ where $\theta_a$ has the same structure as the model $\theta$
 \FOR{$t = 1, \ldots, T$}
 \STATE $S_t \leftarrow$ Sample a set of users with a probability $q$
 \FOR{each legitimate client $i \in S^t \setminus \mathcal{C}$}
 \STATE set $\theta_i^t = \theta^t$ \# where $\theta^1$ is randomly initialized
    \FOR{$k = 1, \ldots, K$}
    \STATE sample mini-batch $B \subset D_i$
    \STATE $\theta^{k+1}_i = \theta^k_i - \gamma \bigtriangledown_{\theta^{k}_i}  L_i (B)$
    \ENDFOR
    \STATE $\bigtriangleup \theta_i^t \leftarrow \theta_i^K -\theta_i^t$
\ENDFOR
\FOR{each compromised client $c \in S^t \cap \mathcal{C}$}
 \STATE $\bigtriangleup \theta_c^t \leftarrow \big(\psi_c^t \sim \mathcal{U}[a, b]\big)\big[ X - \theta^t \big]$
\ENDFOR
    \STATE $\theta^{t+1} \leftarrow \theta^t - \lambda \big[\sum_{i\in S^t \setminus \mathcal{C}  } \bigtriangleup \theta_i^t + \sum_{c\in S^t \cap \mathcal{C}  } \bigtriangleup \theta_c^t \big]   / |S^t|
     $
 \ENDFOR
\end{algorithmic} 
\end{algorithm}


To increase stealthiness of our attack, we introduce a dynamic learning rate $\psi_c^t$, randomly sampled from a predetermined distribution, such as $\mathcal{U}[a, b]$ ($0 < a < b \leq 1$). Before sending the malicious gradients to the server in each training round $t$, they are multiplied by the sampled dynamic learning rate $\psi_c^t$, as follow:

\begin{equation}
\forall c \in \mathcal{C}, t \in [T]: \bigtriangleup \theta_c^t = \psi_c^t \big[X - \theta^t\big].
\label{malicious gradients}
\end{equation}


\begin{figure*}[t]
\begin{minipage}{.69\textwidth}
\hspace{1cm}
\subfigure[Benign clients in normal  FL training and Compromised clients in  \textsc{CollaPois}]{\includegraphics[scale=0.2]{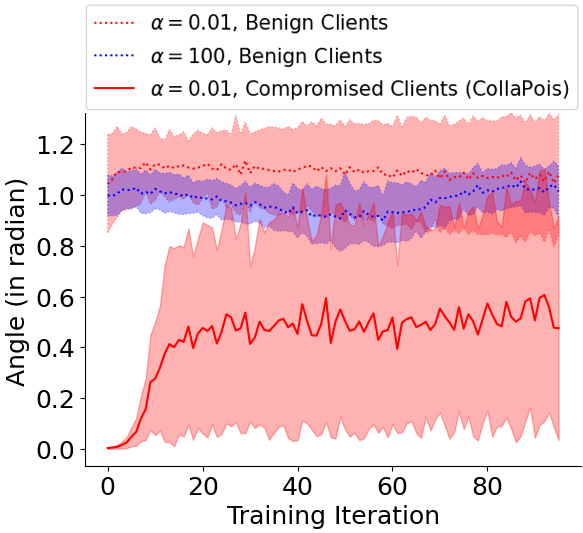}} \vspace{-2.5pt} \hspace{1cm}
\subfigure[Compromised clients in \textsc{DPois} and \textsc{CollaPois}]{\includegraphics[scale=0.3]{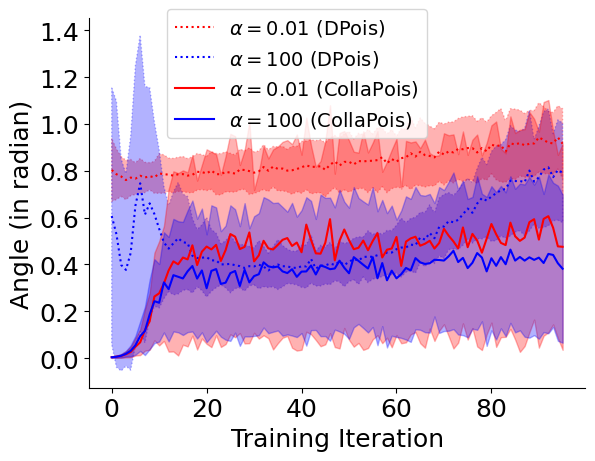}}  \vspace{-2.5pt}
  \captionof{figure}{Average angles among  gradients from benign and compromised clients as a function of $\alpha$ in the FEMNIST dataset. Model and data configuration are in Section \ref{Experimental Results}. }
   \vspace{-5pt}
  \label{visualization-angle}
\end{minipage}%
\hspace{0.1cm}
\begin{minipage}{.3\textwidth}
  \centering
  \includegraphics[scale=0.28]{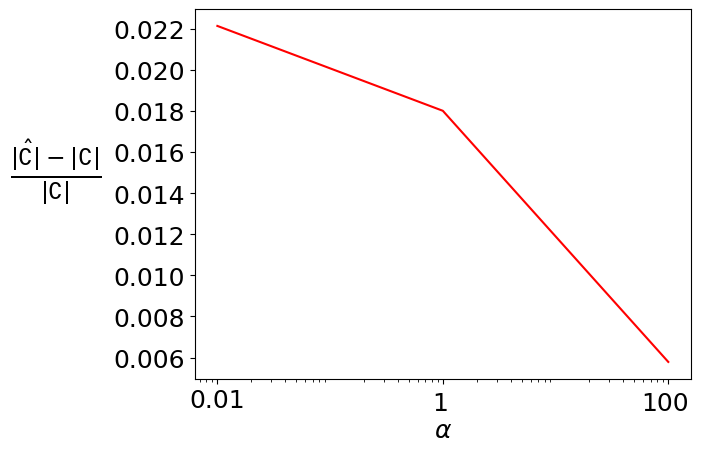}
  \captionof{figure}{Approximation error for the lower bound of $| \mathcal{C}|$ in Theorem \ref{Theorem-angles} as a function of $\alpha$ using  FEMNIST dataset.}
  \vspace{-5pt}
  \label{Capproximate}
\end{minipage}
\end{figure*} \setlength{\textfloatsep}{2pt}
 \vspace{-5pt}

\textbf{Novelty and Advantages.} The novelty and distinctive advantages of \textsc{CollaPois} stem from its simplicity and efficiency. Its simplicity lies in the fact that the attacker only needs to train the Trojaned model $X$ with minimal adjustments to the FL procedure compared to classical data poisoning methods \cite{suciu2018does,li2016data}. Additionally, these adjustments do not add any extra computational overhead, as compromised clients do not need to derive gradients from their local data in each training round. Importantly, the attacker does not require additional knowledge from benign clients or the server to implement this adjustment.

\textsc{CollaPois} offers significant cost-effectiveness for compromised clients when computing malicious gradients using Eq. \ref{malicious gradients} compared to  conventional data poisoning approaches, where local models are trained on poisoned datasets.  
It also benefits the attacker controlling all compromised clients. 

Consequently, \textsc{CollaPois} is a practical and feasible method for uncovering backdoor risks in FL, thanks to its simplicity, efficiency, and the aforementioned benefits.

The following analysis provides more insights into the novelty and key advantages of \textsc{CollaPois}.


\subsection{Smaller and Bounded Numbers of Compromised Clients} 

Given scattered gradients from benign clients, the effectiveness of the malicious local gradients $\bigtriangleup \theta_c^t$ in transferring backdoors to FL models is enhanced.  This allows establishing a lower bound on the number of compromised clients needed for \textsc{CollaPois}, reducing the attack's client requirements.



In Fig.~\ref{visualization-angle}, we present a visual representation to aid our comprehension of the scatter observed among the gradients of both benign and compromised clients. This scatter is illustrated by the angles formed among these gradients. The diversity in clients' local data distribution is expressed through smaller values of the concentration parameter $\alpha$ in the Dirichlet distribution among clients' local data. As a result, angles between pairs of benign clients' local gradients become larger, indicating a more scattered distribution.

This observation can be easily understood, as the local models $\{\theta^t_{i}\}_{i \in N \setminus \mathcal{C}}$ of benign clients are customized through training on their respective local datasets $\{D_i\}_{i \in N \setminus \mathcal{C}}$. When these datasets  $\{D_i\}_{i \in N \setminus \mathcal{C}}$ exhibit greater diversity, the resulting local models become more dispersed. Consequently, their corresponding local gradients $\{\bigtriangleup \theta_i^t = \theta_i^t - \theta^t\}_{i \in N \setminus \mathcal{C}}$ experience more scattering when compared to the same global model $\theta^t$. This, in turn, weakens the aggregation of benign gradients, denoted by $\sum_{i \in S^t \setminus \mathcal{C}} \bigtriangleup \theta^{t}_i$ in Eq. \ref{PoisFedAvg}, in the face of the poisoned gradients $\sum_{c \in S^t \cap \mathcal{C}} \bigtriangleup \theta^{t}_c$. This observation applies to various training algorithms, including FedAvg and FedDC (personalized federated training approaches).

In typical \textsc{DPois} attacks (as illustrated in Fig.~\ref{visualization-angle}b), the malicious local gradients $\{\bigtriangleup \theta_c^t = \theta_c^t - \theta^t\}_{c \in \mathcal{C}}$, where $\{\theta^{t}_{c} = \arg\min_{\theta^t} L(\theta^t, D_c \cup D^{Troj}_c)\}_{c \in \mathcal{C}}$, exhibit a similar level of scatter as benign gradients. This is because the local Trojaned models $\{\theta^{t}_{c}\}_{c \in \mathcal{C}}$ heavily rely on diverse local data distributions, causing them to scatter in each training round.
As local data diversity among compromised clients rises, the angles between malicious gradients increase (Fig.~\ref{visualization-angle}b), reducing \textsc{DPois} attack effectiveness and limiting insights into the impact of data distribution on attack stealthiness.


In this work, we leverage our observations about the gradient scatter to establish a novel correlation between data distribution and the effectiveness as well as the stealthiness of the attack.
We utilize the scatter observed in benign gradients to manipulate the angles among malicious gradients (as illustrated in Fig.~\ref{visualization-angle}a). By consistently reinforcing the aggregation of these malicious gradients during training iterations, we exert a pulling force on the global model $\theta^t$, steering it towards the shared Trojaned model $X$. As a result, we are able to substantially reduce the number of compromised clients needed to successfully execute a backdoor poisoning attack and establish a minimum threshold for the same. We consider a poisoning attempt to be successful in a particular training round $t$ if the updated global model $\theta^{t+1}$ moves closer to the Trojaned model $X$. In other words, malicious gradients dominate benign ones, causing the global model to align with their direction and trigger the backdoor attack. 


In the worst case, when the aggregated benign gradients are oriented in the opposite direction to that of the aggregated compromised gradients, and the angle $\beta_i$ between the gradients of an arbitrary benign client $i$ and the aggregated malicious gradients of all compromised clients follows a normal distribution $\mathcal{N}(\mu_{\alpha}, \sigma^2)$ (Fig.~\ref{visualization-angle}), we derive a lower bound on the minimum number of compromised clients $|\mathcal{C}|$ needed for attack success in a single training round in the following theorem.

\begin{theorem}
The minimum number of compromised clients needed to carry out backdoor poisoning successfully  
in the worst-case scenario is given by the following formula: \vspace{-2.5pt} 
\begin{small}
\begin{equation}
|\mathcal{C}| \ge \frac{2-\sigma^2 - \mu_{\alpha}^2}{ a+b +2-\sigma^2 - \mu_{\alpha}^2} |N|,
\label{Lowerbound C} \vspace{-3pt}  
\end{equation} 
\end{small}

\noindent where $\beta_i$ is the angle between the gradients of an arbitrary benign client $i$ and that of the aggregated malicious gradients of all the compromised clients. We assume that $\beta_i$ follows a normal distribution, i.e., $\beta_i \sim \mathcal{N}(\mu_{\alpha}, \sigma^2)$. 
\label{Theorem-angles}
\end{theorem}

\textbf{All proofs of Theorems are in Supplementary.}

Theorem \ref{Theorem-angles} is obtained by approximating $\sum_{c \in \mathcal{C}}  \psi_c $ with  $|\mathcal{C}| \cdot \frac{ (a+b)}{2}$  and  $\sum_{i \in N\backslash\mathcal{C}} \beta_{i}^{2}$ with  $\mathbb{E}(\sum_{i \in N\backslash\mathcal{C}} \beta_{i}^{2})$. The error of the approximation can be bounded using concentration bounds, such as Hoeffding, with high confidence levels.

In practical scenarios, the attacker can make accurate estimations of the mean $\mu_\alpha$ and variance $\sigma$ of the angles by utilizing the datasets $\{D_c\}_{c \in \mathcal{C}}$ collected by compromised clients. This enables the attacker to precisely approximate the lower bound of $| \mathcal{C} |$ with a bounded error based on concentration bounds, such as Hoeffding bound. Fig.~\ref{Capproximate} presents this relative approximation error $| \frac{|\hat{\mathcal{C}}| - |\mathcal{C}|}{|\mathcal{C}|} |$ as a function of the concentration parameter $\alpha$, where $|\hat{\mathcal{C}}|$ is the approximated lower bound of $|\mathcal{C}|$. The higher the degree of diversity in benign clients' local data is, the larger the relative approximation error is. However, the relative approximation error is marginal across all the degrees of $\alpha$, i.e., 2.23\% given $\alpha=0.01$ and 0.57\% given $\alpha=100$. In addition, the mean of angles $\mu_\alpha$ and its variance $\sigma$ are quite consistent from initial training rounds and throughout the training process 
(Fig.~\ref{visualization-angle}); therefore, the attacker can estimate the lower bound $|\mathcal{C}|$ in less than ten training rounds. This reduces poisoning delay in federated training. Importantly, our lower bound of $|\mathcal{C}|$ remains practical,  as the attacker follows the threat model without extra client information.

\begin{figure}[t] \vspace{-10pt}
      \centering 
       \includegraphics[scale=0.15]{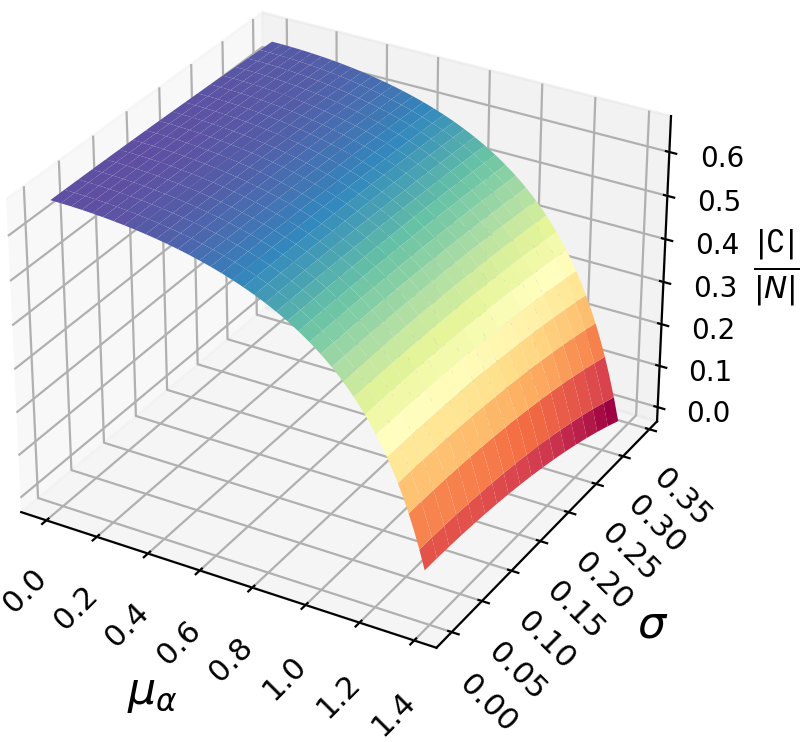}  \vspace{-5pt}
      \caption{3D plot of $|\mathcal{C}| / |N|$ as a function of $\mu_\alpha$ and $\sigma$. }
      \label{3d lower bound}
\end{figure}

Fig.~\ref{3d lower bound} shows the lower bound of $\frac{|\mathcal{C}|}{|N|}$ in a 3D surface as a function of $\mu_\alpha$ and $\sigma$. From Theorem \ref{Theorem-angles}, a higher mean $\mu_\alpha$ and variance $\sigma$   (indicating more scattered gradients and greater local data diversity) reduce the number of compromised clients $\mathcal{C}$ needed for successful execution of \textsc{CollaPois}, leading to a higher backdoor success rate with more diverse local data. 


To address this concern, \textsc{CollaPois} introduces the concept of a shared Trojaned model $X$ as a stable and optimized poisoned area. Leveraging the lower bound on the number of compromised clients, we demonstrate in the following theorem that the global FL model $\theta$ converges to a small bounded region around  $X$. This ensures that the impact of the backdoor attack is confined to a limited area.
\begin{theorem}
For a compromised client $c$ participating in round $t$, the $l_2$-norm distance between the global model $\theta^t$ and the Trojaned model $X$ is always bounded as follows:
\begin{equation}
\|\theta^t - X\|_2 \leq (1/a - 1) \| \bigtriangleup \theta_c^{t'} \|_2 + \|\zeta\|_2,
\end{equation}
where $\forall t: \psi^t_c \sim \mathcal{U}[a, b]$,   $a < b$, $a, b \in (0,1]$,
$t'$ is the closest round the compromised client $c$ participated in, and $\zeta$ 
is a small error rate.
\label{ConvergenceBound}
\end{theorem}

In Theorem \ref{ConvergenceBound}, as the global model approaches convergence, indicated by $t'$ and $t$ approaching the number of rounds $T$, the norm $\|\xi\|_2$ becomes extremely small, and $\| \bigtriangleup \theta_c^{t'} \|_2$ is bounded by a small constant $\tau$. This ensures that the global FL model $\theta^T$ converges to a bounded and low-loss region surrounding the Trojaned model $X$. In other words, $\|\theta^T - X\|_2$ is minimized to a negligible value. This provides a robust assurance of the success of our attack.

Theorem \ref{ConvergenceBound} shows that the $l_2$-norm distance between the global model $\theta^t$  and the Trojaned model $X$, i.e., $\|\theta^t - X\|_2$, is bounded by $(\frac{1}{a} - 1) \| \bigtriangleup \bar{\theta}_c^{t'} \|_2 + \|\xi\|_2$, where $t'$ is the closest round the compromised client $c$ participated in, and $\xi \in \mathcal{R}^{m}$ is a small error rate. When the FL model converges, e.g., $t', t \approx T$, $\|\xi\|_2$ become tiny and $\| \bigtriangleup \bar{\theta}^c_{t'} \|_2$ is bounded by a small constant $\tau$ ensuring that the output of the FL model under \textsc{CollaPois} given the compromised client $\theta^T_c$ converges into a bounded and low loss area surrounding 
$X$ ($\|\theta^T_c - X\|_2$ is tiny) to imitate the model convergence behavior of legitimate clients.
 Consequently, \textsc{CollaPois} requires 
 a small number of compromised clients to be highly effective. Also, \textsc{CollaPois} is stealthy by avoiding degradation and shifts in model utility on legitimate data samples during the whole poisoning process.
 

\vspace{5pt}
\noindent\fbox{%
    \begin{varwidth}{0.465\textwidth}
\textbf{Remark.} Theorems \ref{Theorem-angles} and \ref{ConvergenceBound} establish that \textsc{CollaPois} maintains stealthiness through   controlled perturbations, with minimal impact on the effectiveness of clean inputs. Simultaneously, it demonstrates high effectiveness even with a limited number of compromised clients, thanks to its well-coordinated malicious updates.
    \end{varwidth}%
} 

\vspace{-1pt}
\subsection{Attack Stealthiness}

In addition to their effectiveness, the malicious gradients $\{\bigtriangleup \theta_c^t\}_{c \in \mathcal{C}}$ possess several key properties that contribute to their stealthiness, including the following:

\textbf{(1)} 
The Trojaned model $X$ exhibits higher model utility on legitimate data samples compared to randomly initialized global FL and benign clients' local models, particularly during the early training rounds and when there is a greater diversity in the local data distribution of benign clients (indicated by smaller values of the concentration parameter $\alpha$). The resulting models achieve superior model utility on legitimate data samples by utilizing malicious gradients to train both the global and clients' local models. Consequently, \textsc{CollaPois} demonstrates greater stealthiness compared to \textsc{MRepl} and \textsc{DPois} attacks, as it avoids the degradation and shifts in model utility on legitimate data samples throughout the entire poisoning process. This characteristic makes the detection of \textsc{CollaPois} highly challenging during the federated training.

\textbf{(2)} By ensuring that the random and dynamic learning rate $\psi_c^t$ is exclusively known to the compromised client $c$, we can effectively prevent the server from tracking the Trojaned model $X$ or detecting suspicious behavior patterns from the compromised client. In practice, the server can identify compromised clients with precision $p$ and detect the presence of the Trojaned model $X$. The server's set of identified compromised clients consists of $p \times |\mathcal{C}|$ compromised clients $\bar{\mathcal{C}}$ and $(1-p) (|N|- |\mathcal{C}|)$ benign clients $\bar{L}$. The estimated Trojaned model is $X' = \sum_{c \in \bar{\mathcal{C}} \cup \bar{L} } \theta^t_c / |\mathcal{C}|$. 
The following theorem establishes a bound on the server's $l_2$-norm estimation error of the Trojaned model $X$, denoted as $Error = \| X' -X\|_2$.
\begin{theorem}
The server's estimation error of the Trojaned model $X$ is bounded as follows:
\begin{small} \vspace{-5pt}
\begin{equation}
  \big\| \sum_{c \in \bar{\mathcal{C}} } \frac{\bigtriangleup \theta^t_c}{p|\mathcal{C}| b } \big\|_2  \le  Error \le \arg\max_{L \subseteq N \text{  s.t. } |L|=|\mathcal{C}| }  \big\| \sum_{i \in L } \frac{\theta^t_i}{|L|} - X \big\|_2.
  \label{errorbound}
\end{equation} \vspace{-10pt}
\end{small}
\label{XapproximationError}
\end{theorem}

\begin{figure}[t]
      \centering
       \includegraphics[scale=0.17]{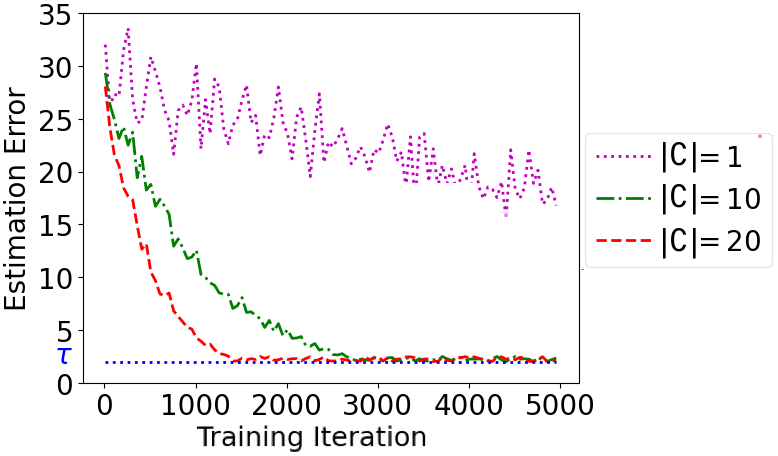}\vspace{-5pt} 
      \caption{Estimation error of \textsc{CollaPois}. ($p = 1$, FEMNIST)} 
      \label{estimation error}
\end{figure}

From Theorem~\ref{XapproximationError}, we observe that: \textbf{(1)}  Lower detection precision $p$ leads to a larger estimation error near the upper bound; \textbf{(2)}  A smaller upper bound $b$ of $\psi_c^t$ increases the estimation error's lower bound; and \textbf{(3)}  If the malicious gradient $\bigtriangleup \theta_c^t$ is too small, we can uniformly upscale its $l_2$-norm to be a small constant, denoted $\tau$, to enlarge the lower bound of the estimation error without affecting the model utility or backdoor success rate. 
 Fig.~\ref{estimation error} shows this effect with $p=1$ across various numbers of compromised clients $|\mathcal{C}|$. After 1,000 rounds, the error stabilizes at a controlled lower bound ($\tau=2$), preventing accurate estimation of model $X$. 


\vspace{5pt}
\noindent\fbox{%
    \begin{varwidth}{0.465\textwidth}
\textbf{Remark.} Practitioners can connect Theorems \ref{Theorem-angles} and \ref{XapproximationError} to discover that: The more diverse clients' local data is (i.e., larger values of $\mu_\alpha$ and $\sigma$ resulting in a smaller number of compromised clients $|\mathcal{C}|$ in Eq. \ref{Lowerbound C}), the more difficult for the server to approximate the Trojaned model $X$ is; hence, the more stealthy the attack will be. This is because a smaller number of compromised clients $|\mathcal{C}|$ induces a larger lower bound of the estimation error in Eq. \ref{errorbound}.
    \end{varwidth}%
} \vspace{2pt}

\begin{figure}[t]
      \centering
       \includegraphics[scale=0.2]{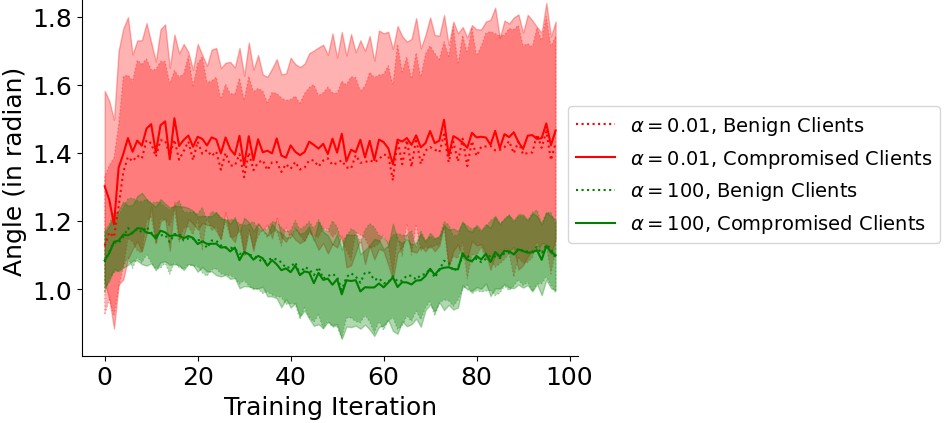} \vspace{-5pt}
      \caption{Attack Stealthiness: Angles between malicious/benign gradients and sampled gradients. Compromised clients with benign clients are blended and modestly different (using the FEMNIST dataset with $\psi_c^{t} \sim \mathcal{U}[0.95, 0.99]$).} \vspace{-5pt}
      \label{Stealthiness Angles}
\end{figure}
\setlength{\textfloatsep}{5pt}

\setlength{\textfloatsep}{2pt}

While the server may not be efficient in directly estimating the Trojaned model $X$, it can attempt to distinguish compromised clients by analyzing the angles and magnitudes of the gradients submitted by each client \cite{10.1145/3576915.3623212}. 

To protect malicious gradients from being detectable, the attacker can marginally adjust the dynamic learning rate $\psi_c^{t}$ to seamlessly blend each malicious gradient in the background of benign gradients. The wider the range of $\psi_c^{t} \sim \mathcal{U}[a, b]$, the more scattered malicious gradients in terms of angles and magnitude are, i.e., more randomness. The attacker can select a suitable range of $\mathcal{U}[a, b]$ such that the average angle and its variance between each of the malicious gradients and a set of sampled gradients (which plays a role of data background) are similar to those of benign clients. 
In practice, the attacker can derive sampled gradients using clean data from compromised clients ${D_c}_{c \in \mathcal{C}}$ and the global model $\theta^t$. These clean gradients can then mimic those from benign clients, ensuring the threat model without accessing additional benign client information.


Fig.~\ref{Stealthiness Angles} shows that malicious and benign gradients have similar average angles and variance. To improve robustness, malicious gradients are clipped with a shared bound $A$, keeping their magnitude within the range of benign gradients.




Consequently, \textsc{CollaPois} can conceal malicious gradients (angles, variance, magnitude) to bypass statistical and clustering defenses \cite{10.1145/3576915.3623212} without compromising attack performance, provided $\psi_c^t \sim \mathcal{U}[a, b]$ and the clipping bound $A$ are chosen such that they do not negatively impact federated training.

\vspace{5pt}
\noindent\fbox{%
    \begin{varwidth}{0.465\textwidth}
\textbf{Remark.}
Our attack has several key advantages: (1) It requires only a small, bounded number of compromised clients to successfully manipulate the global model toward a tight region around the Trojaned model $X$ (Theorems \ref{Theorem-angles}, \ref{ConvergenceBound}), enabling effective backdoor transfer to local models. (2) It achieves higher attack success rates under diverse local data distributions, enhancing real-world applicability (Theorem \ref{Theorem-angles}). (3) It prevents accurate estimation of $X$ or detection of malicious clients (Theorem \ref{XapproximationError}), ensuring stealth.  Overall, these advantages collectively contribute to the effectiveness, practicality, and stealthiness of our attack against FL systems.
    \end{varwidth}%
} 

\section{Experimental Results}
\label{Experimental Results}

In this section, we seek to 
examine the connections among backdoor attacks, defense mechanisms, different local data distribution levels, and the performance of Federated Learning (FL). To achieve this, we focus 
on addressing four demanding inquiries:
\textbf{(1)} How effective is \textsc{CollaPois} as a poisoning technique in FL, compared to existing backdoor attacks? \textbf{(2)} How does \textsc{CollaPois} perform with different levels of data diversity given different proportions of compromised clients? \textbf{(3)} How to defend against \textsc{CollaPois}, and what are the costs and limitations of such defenses? and \textbf{(4)} How many and which clients are affected, at which attack success rates, and why?


To answer these questions, we evaluate  at the \textit{\textbf{population level}}, considering the FL system as a whole, and at the \textit{\textbf{client level}} to study the impact of the attack on each client. 

\textbf{Data and Model Configuration.} 
We conduct experiments on  Sentiment \cite{go2009twitter} and FEMNIST  \cite{caldas2018leaf} datasets.  
We leverage the symmetrical Dirichlet distribution with different values of the concentration parameter $\alpha \in [0.01, 100]$ \cite{gao2022feddc}, where smaller $\alpha$ indicates greater diversity. 
In Sentiment, we include $5,600$ clients with over $1$ million   samples. In FEMNIST, there are $3,400$ clients with $805,263$ samples. We use   $q=1\%$ and $\psi \sim \mathcal{U}[0.9, 1]$. Class 0 is designated as the target class. Data is split into 70\% training, 15\% testing, and 15\% validation per client. Combined validation sets from all compromised clients form the auxiliary set to train the Trojaned model $X$. The attacker randomly compromises 0.1\%, 0.5\%, and 1\% of clients, treating these small fractions as a practical threat \cite{drawingboard2022}.



We adopt the model from \cite{shamsian2021personalized} with a LeNet-based network for the local model and a fully connected network with  linear heads for the global model. For the Sentiment dataset, we use the BERT tokenizer with a two-layer fully connected task head. The SGD optimizer is applied with a learning rate of $0.01$ for the global model and $0.001$ for local models.

\textbf{Evaluation Approach.} We evaluate \textsc{CollaPois} via   three approaches. We first compare \textsc{CollaPois} with \textsc{DPois}, \textsc{MRepl}, and distributed backdoor attacks (\textsc{DBA}) \cite{bagdasaryan2020backdoor,xie2020dba} in terms of benign accuracy (Benign AC) on legitimate data samples and backdoor success rate (Attack SR) on Trojaned data samples without defense. Then, we investigate the effectiveness of adapted robust federated training algorithms under a variety of hyper-parameter settings against \textsc{CollaPois}. Finally, we provide a performance summary of the state-of-the-art attacks and defenses to assess the landscape of backdoor risks in FL under diverse levels of clients' local data.

The average Benign AC and Attack SR across all the clients, using testing data, are defined as:

\vspace{-5pt}
{\small
\begin{align}
& \text{Benign AC} = \frac{1}{|N|} \sum_{i \in N} \Big[ \frac{1}{|D_i^{test}|} \sum_{x \in D_i^{test}} \mathbb{I} \big(f(x,\theta_i), y \big) \Big] \nonumber \\
& \text{Attack SR} =  \frac{1}{|N|} \sum_{i \in N} \Big[ \frac{1}{|D_i^{test}|} \sum_{x \in D_i^{test}} \mathbb{I} \big(f(x + \mathcal{T},\theta_i), y^{Troj} \big) \Big],  \nonumber
\end{align}
}
\vspace{-5pt}

\noindent where $x + \mathcal{T}$ is a Trojaned data sample, $\mathbb{I}$ is the indicator function s.t. $\mathbb{I}(y', y) = 1$ if $y' = y$; otherwise $\mathbb{I}(y', y) = 0$, and $D_i^{test}$ and $|D_i^{test}|$ are a test set and its number of samples. 

We evaluate \textsc{CollaPois} against state-of-the-art personalized FL algorithms FedDC \cite{gao2022feddc} and MetaFed \cite{chen2023metafed}, as well as the widely used FedAvg, to assess the attack's generalizability. To analyze client-specific performance, stealthiness, and backdoor attack risks, we report Benign AC and Attack SR values for the top-k\% affected benign clients $i$, selected based on the highest sum  of Benign AC and Attack SR, as follows:

\begin{small} \vspace{-5pt}
\begin{align}
    \text{score}_i = \frac{\sum_{x \in D_i^{test}} \Big[ \mathbb{I} \big(f(x,\theta_i), y \big) + \mathbb{I} \big(f(x + \mathcal{T},\theta_i), y^{Troj} \big) \Big]}{|D_i^{test}|}.
    \label{score}
\end{align} \par \vspace{-5pt}
\end{small}

\textit{\textbf{Figures \ref{cifar10-vis}-\ref{fig:all defenses-femnist-0.1-0.5-mainbody} are in the Supplementary$^1$.}}

\begin{figure}[t] 
\captionsetup[subfigure]{justification=centering}
    \centering
  \subfigure[FedAvg-BenignAC]{\includegraphics[scale=0.25]{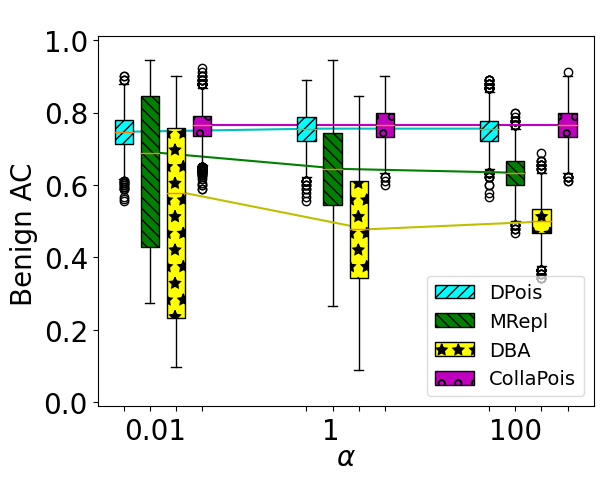}}%
      \vspace{-5pt} \hfill
      \subfigure[FedAvg-AttackSR] {\includegraphics[scale=0.25]{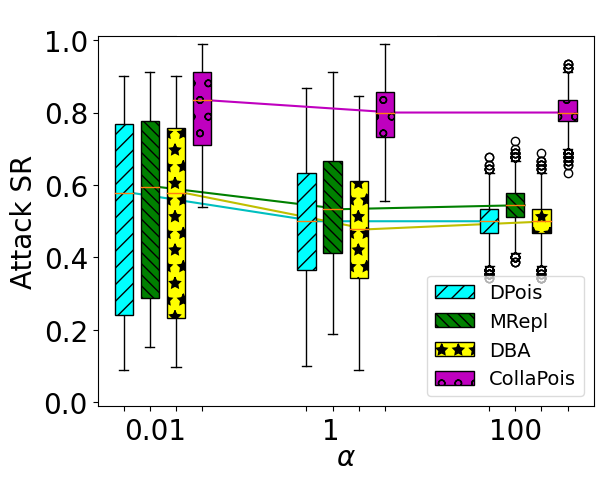}} \hfill \vspace{-5pt}
     \subfigure[FedDC-BenignAC] {\includegraphics[scale=0.25]{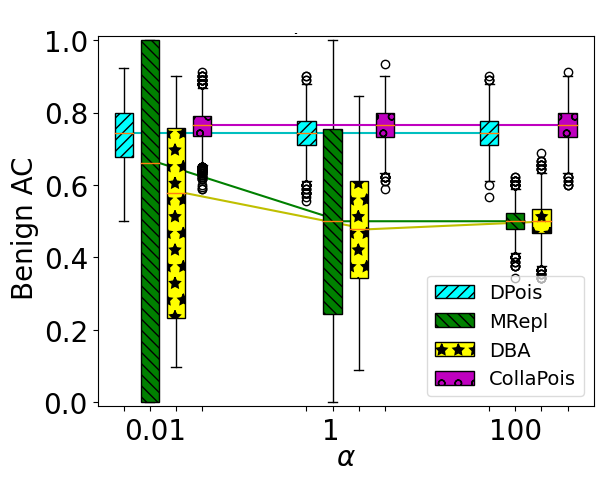}}  \vspace{-5pt}\hfill
     \subfigure[FedDC-AttackSR] {\includegraphics[scale=0.25]{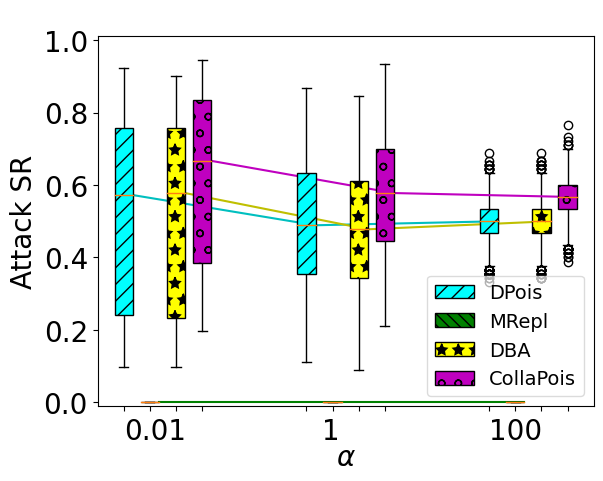} }\hfill \vspace{-5pt}
    \subfigure[MetaFed-BenignAC] 
       {\includegraphics[scale=0.25]{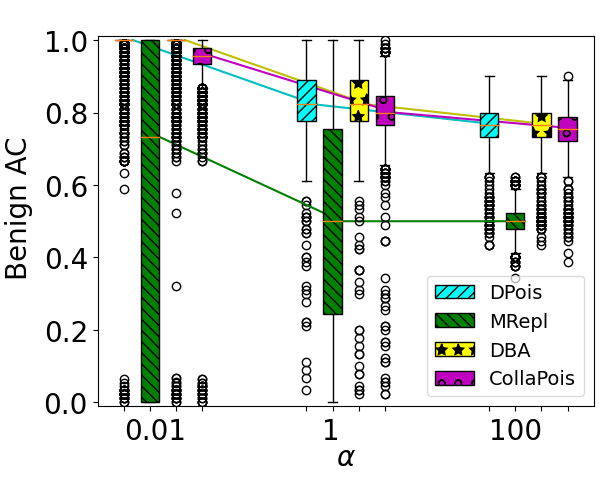}}\vspace{-5pt} \hfill
        \subfigure[MetaFed-AttackSR]
     { \includegraphics[scale=0.25]{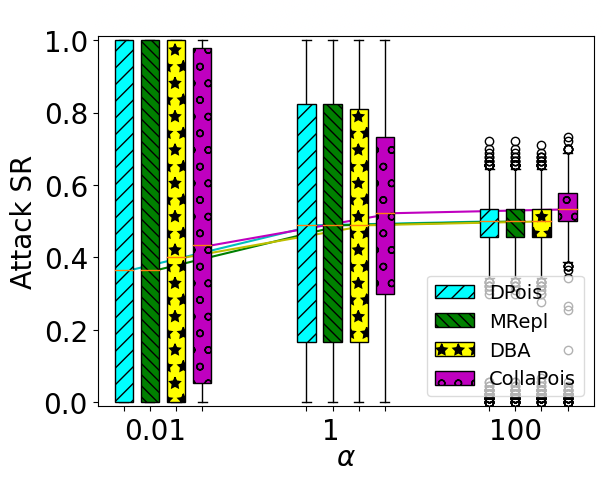}}\vspace{-5pt} 
    \caption{FedAvg, FedDC, and MetaFed under attacks ($1\%$ compromised clients) in the Sentiment dataset. }
    \label{fig:all attacks-sentiment-mainbody}  
\end{figure}

\textbf{\textsc{CollaPois} and Existing Attacks.} 
Figs.~\ref{fig:all attacks-sentiment-mainbody} and \ref{fig:all attacks-femnist-mainbody} illustrate the Benign AC and Attack SR of \textsc{CollaPois} and three baseline poisoning attacks as a function of the concentration parameter $\alpha$ on the Sentiment and FEMNIST datasets across  FL algorithms, where the attacker compromises 1\% of clients.

The figures show that \textsc{CollaPois} significantly outperforms \textsc{DPois}, \textsc{MRepl}, and \textsc{DBA} in Attack SR without a notable drop in Benign AC across datasets, FL algorithms, and  $\alpha$ values. In the Sentiment dataset, \textsc{CollaPois} achieves an 25.56\% increase in Attack SR with a slightly better Benign AC (1.94\%) compared to the best baseline, \textsc{DPois} on average ($p$-value 3.08e-19). All statistical tests are 2-tail t-tests. 
Each experiment was run 5 times with small variance (0.01\%-0.03\%). In the FEMNIST dataset, Attack SR rises to 91.25\% on average ($p$-value 7.78e-167), while baseline attacks struggle under FedDC.
This is due to FedDC's local personalization.  When Trojans are poorly integrated with global and local models during training, local personalization can mitigate backdoor attacks. \textsc{CollaPois} tackles this by aligning global and local models near the Trojaned model $X$. Unlike \textsc{MRepl}, \textsc{DPois}, and \textsc{DBA}, where no such region exists, local personalization struggles to pull the model away from this area.


\textbf{Local Data Diversity and Attack SR.} Figs.~\ref{fig:all attacks-sentiment-mainbody} and \ref{fig:all attacks-femnist-mainbody} show that as $\alpha$ decreases (indicating more diverse local data), the average Attack SR increases. At $\alpha = 0.01$, \textsc{CollaPois} achieves 83.33\% Attack SR, dropping to 80.00\% at $\alpha = 1$ and 79.89\% at $\alpha = 100$.
This observation aligns with our theoretical analysis. A slight difference is observed with MetaFed, as Attack SR shows a minor increase with higher $\alpha$ values.
This result is because MetaFed creates personalized models via knowledge distillation by leveraging common knowledge from neighboring clients. In highly non-IID scenarios, these neighbors are sparse, limiting knowledge transfer and reducing the backdoor's ability to spread beyond compromised clients.


\setlength{\textfloatsep}{2pt}

\begin{figure}[t]
    \centering
 \subfigure[FedAvg-BenignAC]{\includegraphics[scale=0.25]{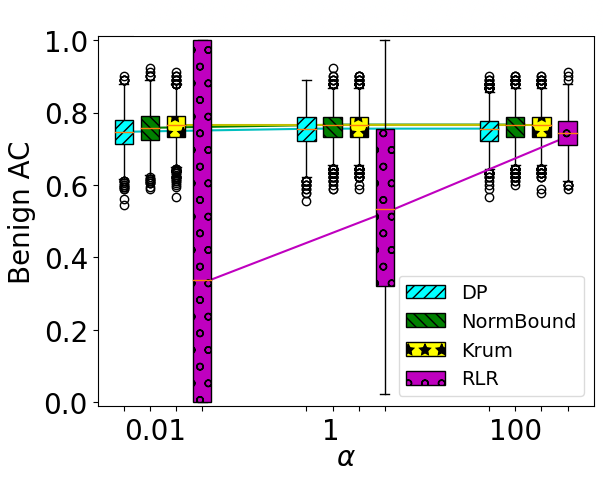}}%
       \vspace{-5pt} \hfill
      \subfigure[FedAvg-AttackSR]{\includegraphics[scale=0.25]{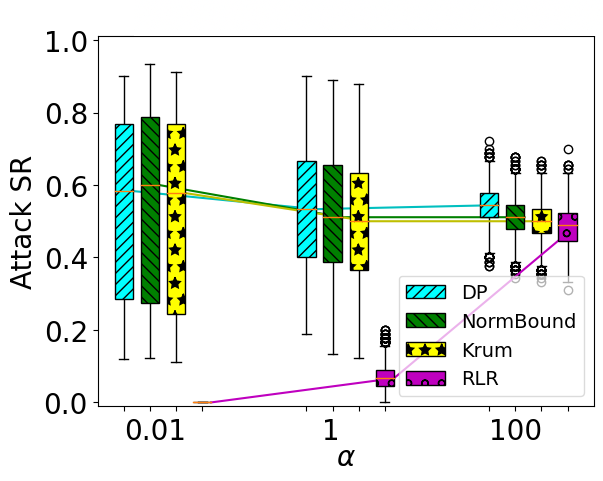}} \vspace{-5pt}
  \subfigure[FedDC-BenignAC]{\includegraphics[scale=0.25]{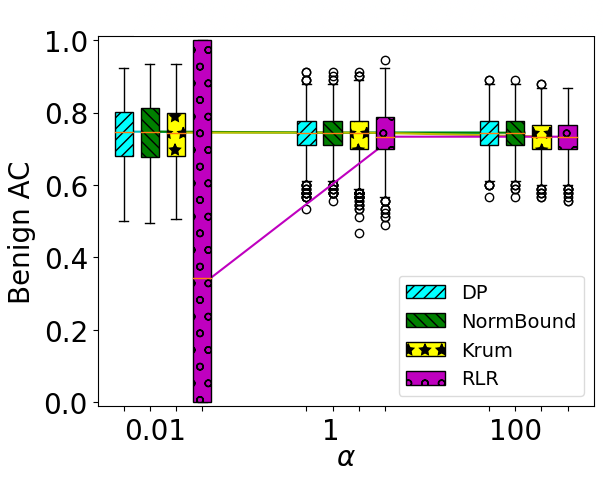}} \vspace{-5pt} \hfill
       \subfigure[FedDC-AttackSR]{\includegraphics[scale=0.25]{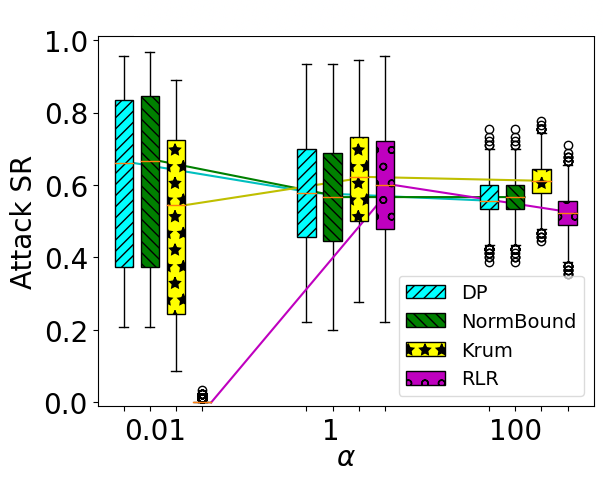}} \vspace{-5pt}
         \subfigure[MetaFed-BenignAC]{
       \includegraphics[scale=0.25]{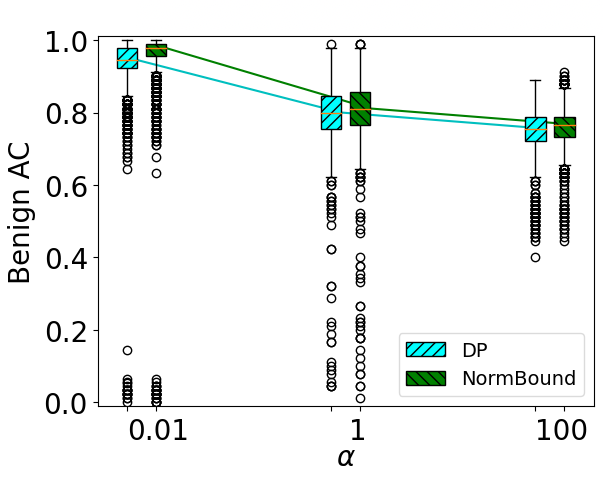}} \vspace{-2.5pt}\hfill
       \subfigure[MetaFed-AttackSR]{\includegraphics[scale=0.25]{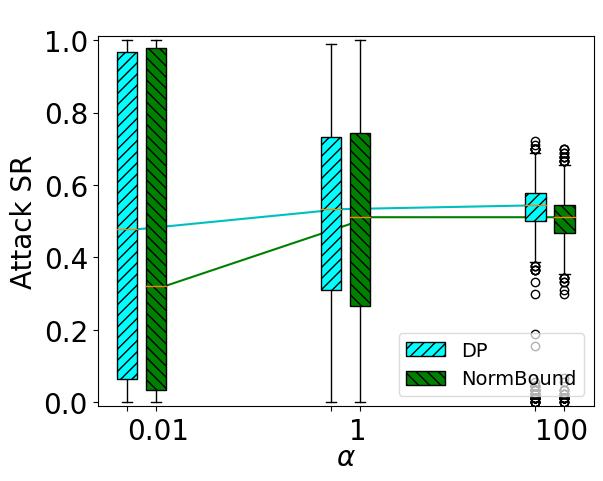} } 
    \caption{\textsc{CollaPois} ($1\%$ compromised clients) under defenses for the Sentiment dataset. (Krum and RLR are not applicable for MetaFed.)}
    \label{fig:all defenses-sent-mainbody}  
\end{figure}\setlength{\textfloatsep}{2pt}


\begin{figure}[t]
    \centering
   \subfigure[FedAvg-BenignAC]{  \includegraphics[scale=0.23]{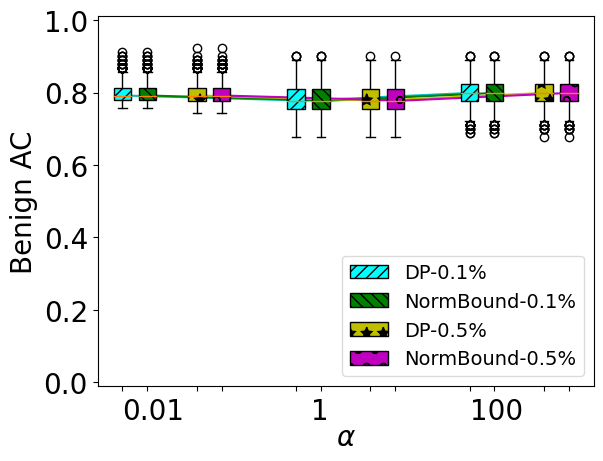}}%
       \vspace{-5pt} \hspace{0.5cm}
     \subfigure[FedAvg-AttackSR]{  \includegraphics[scale=0.23]{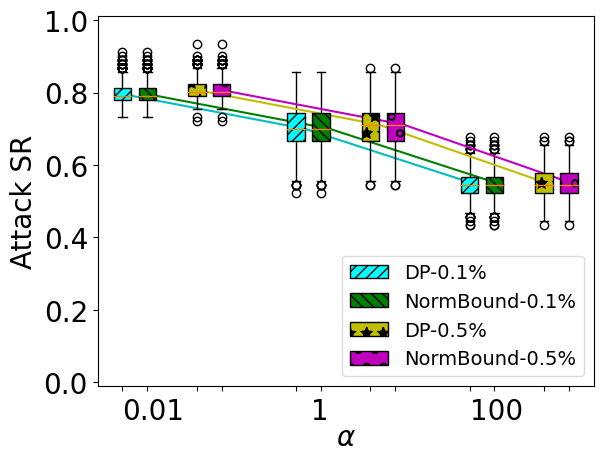}} \vspace{-5pt}
      \subfigure[FedDC-BenignAC]{  \includegraphics[scale=0.23]{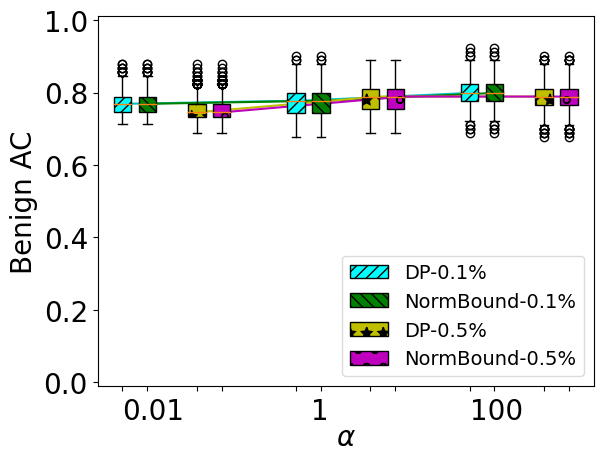}} \vspace{-5pt} \hspace{0.5cm}
      \subfigure[FedDC-AttackSR]{ \includegraphics[scale=0.23]{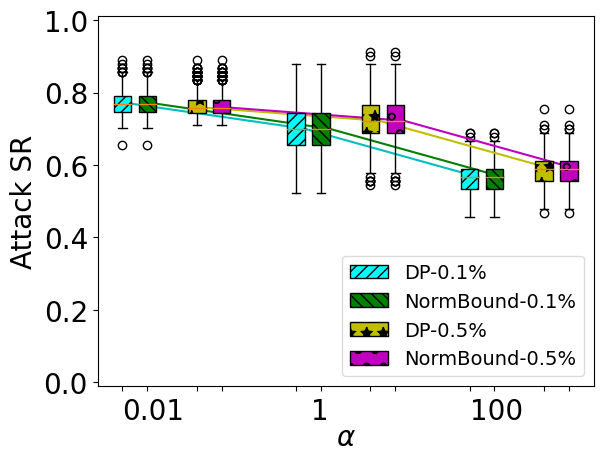}} \vspace{-5pt}
       \subfigure[MetaFed-BenignAC]{ \includegraphics[scale=0.23]{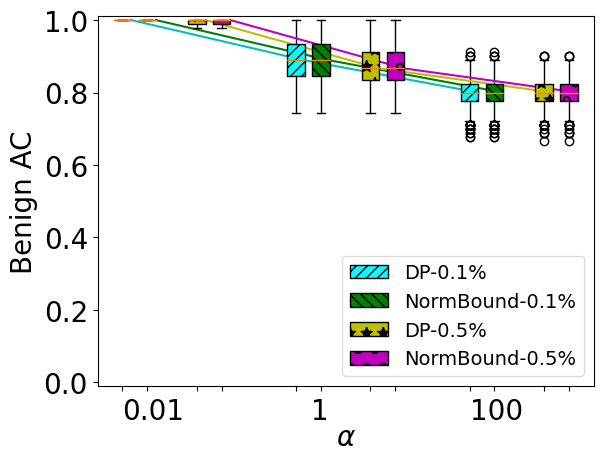}} \vspace{-5pt} \hspace{0.5cm}
    \subfigure[MetaFed-AttackSR]{   \includegraphics[scale=0.23]{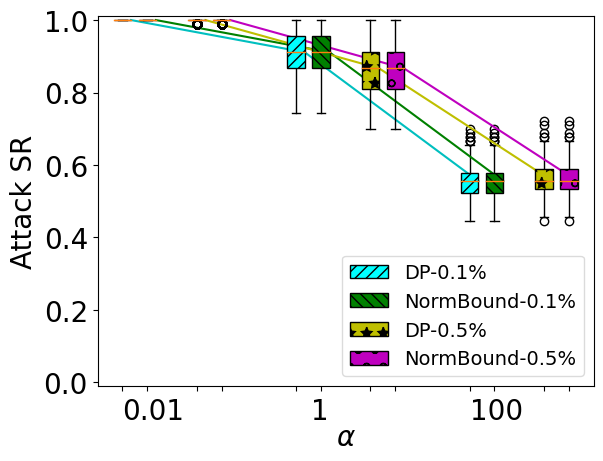}}  
    \caption{ \textsc{CollaPois} (under defenses) with 0.1\% and 0.5\%  compromised clients  for the Sentiment.  (Top 25\% Clients) } 
    \label{fig:all defenses-sent-0.1-0.5-mainbody}  
\end{figure}\setlength{\textfloatsep}{2pt}

Furthermore, smaller $\alpha$ values lead to greater variation in Attack SR across federated training algorithms and datasets due to more diverse clients' local data, resulting in dispersed distributions. Consequently, a subset of benign clients aligns closely with compromised clients and the Trojan model $X$, leading to higher Attack SR, while others are more isolated, showing lower rates. This causes a broader range of Attack SR values. In addition, weaker benign gradients at smaller $\alpha$ values allow the backdoor to infect more benign clients, supported by both average Attack SR  and theoretical analysis.


\textbf{Bypassing Robust Federated Training.} 
Since \textsc{CollaPois} outperforms baseline attacks, we evaluate its effectiveness against robust federated training algorithms. We select four state-of-the-art (SOTA) methods: DP-optimizer (\textbf{DP}), \textbf{NormBound}, \textbf{Krum}, and robust learning rate (\textbf{RLR}). These choices encompass diverse design strategies, allowing a comprehensive evaluation of \textsc{CollaPois} against SOTA defenses.


An effective backdoor defense should maintain a minimal drop in Benign AC while reducing Attack SR (lower is better), allowing efficient FL training while mitigating backdoor  effects. However, Figs.~\ref{fig:all defenses-sent-mainbody} and \ref{fig:all defenses-femnist-mainbody} show the lack of such effective defense among the baseline robust federated training methods.

Standard defenses, i.e., Krum and RLR often lead to substantial drops in Benign AC, making them effective but impractical. Some defenses, such as DP and NormBound, leave FL models highly vulnerable, with Attack SRs as high as 89.02\% and 91.60\%, respectively. Krum and RLR reduce Benign AC by 24.93\% and 61.53\% on average ($p$-value 8.07e-9). Only MetaFed combined with DP or NormBound on FEMNIST shows promise, maintaining high Benign AC and lower Attack SR. However, even then, over 60\% of benign clients are compromised across various $\alpha$ levels (Fig.~\ref{fig:all defenses-femnist-mainbody}f).

\textbf{Bypassing Defenses.} 
\textsc{CollaPois}  can hide the malicious gradients (i.e., in terms of angles, variance, and magnitude)  bypassing the SOTA statistical tests and clustering-based defenses \cite{10.1145/3576915.3623212} without performance loss. 
There is no significant difference between malicious and benign gradients using t-test for the average angle and mean, Levene's test \cite{LIM1996287} for the variances, Kolmogorow-Smirnow-Test \cite{16e7f618-c06b-3d10-8705-1086b218d827} for the gradients' distributions, and only a tiny 3.5\% chance that a malicious gradient is disregarded as an outlier based on the 3$\sigma$ rule \cite{43a4cc54-e2d7-3026-8ece-ba4cf618091a}.

\textbf{Percentage of Compromised Clients.} To identify when a defense becomes effective against \textsc{CollaPois}, we reduce compromised clients from 1\% to 0.5\% and 0.1\%, indicating very small numbers of compromised clients, 5 and 28 clients in the Sentiment dataset, and 4 and 7 clients in the FEMNIST dataset.
Lower Attack SR with high average Benign AC across clients is expected (Figs.~\ref{fig:all defenses-sent-app-0.1}-\ref{fig:all defenses-femnist-app-0.5}), but this does not indicate effective defense.
The top-25\% of infected benign clients  show very high Attack SR (86.12\% on average with 0.5\% compromised clients) across datasets and robust federated training algorithms 
(Figs.~\ref{fig:all defenses-sent-0.1-0.5-mainbody} and \ref{fig:all defenses-femnist-0.1-0.5-mainbody}). The Attack SR is even higher for top-1\% infected clients (Figs.~\ref{fig:all defenses-sent-app-0.5-top1} and \ref{fig:all defenses-femnist-app-0.5-top1}). Also, we observe high Attack SR for top-50\% infected clients  (Figs.~\ref{fig:all defenses-sent-app-0.5-top50} and \ref{fig:all defenses-femnist-app-0.5-top50}). 
While generally effective with 0.1\% compromised clients (74.65\% Attack SR on average), defenses like FedAvg with DP or NormBound show promising results in the FEMNIST dataset (Fig.~\ref{fig:all defenses-femnist-0.1-0.5-mainbody}a), achieving a low 4.55\% Attack SR with a 23.83\% Benign AC drop. 
Hence, even a small fraction of compromised clients (0.1-0.5\%) allows \textsc{CollaPois} to compromise a significantly large portion (25\%) of benign clients, with an average Attack SR over 60.12\%. At 1\% compromised clients, all benign clients are  affected.

\textbf{Client-level Evaluation.} Our  results reveal that different clients exhibit varying levels of backdoor susceptibility, as indicated by the spectrum of Attack SR  resulting from \textsc{CollaPois}, raising a fundamental question: What underlies these discrepancies in backdoor risk among benign clients?

\begin{figure}[t]
      \centering
       \includegraphics[scale=0.23]{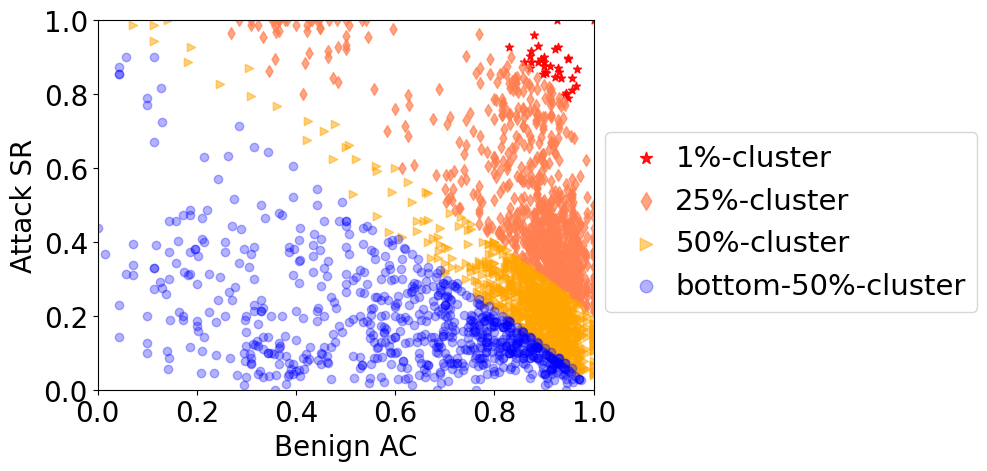} \vspace{-5pt}
      \caption{Benign AC and Attack SR for all clients in the FEMNIST dataset using FedAvg under DP defense.} \vspace{-5pt}
      \label{client-level-mainbody}
\end{figure}
\setlength{\textfloatsep}{5pt}

\begin{figure}[t]
  \centering
\subfigure[FEMNIST]{\includegraphics[scale=0.152]{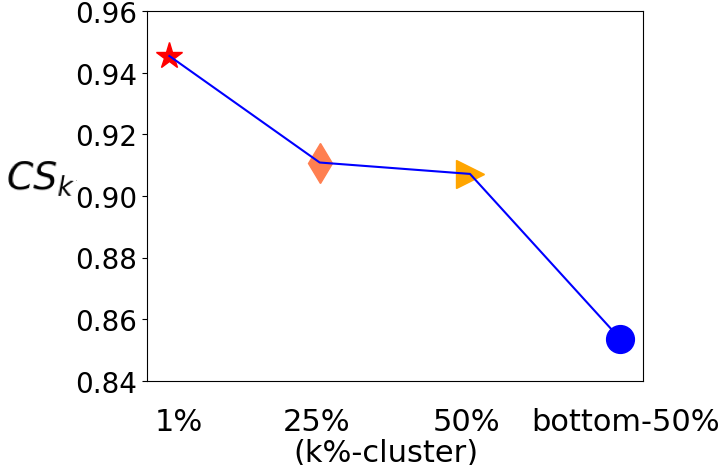}}\hspace{0cm}
\subfigure[Sentiment]{\includegraphics[scale=0.2]{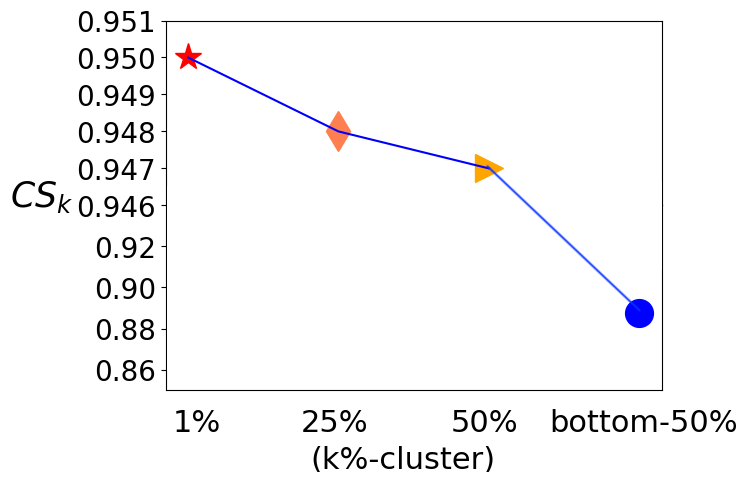}} \vspace{-10pt} \hspace{0cm} 
\subfigure[FEMNIST]{\includegraphics[scale=0.16]{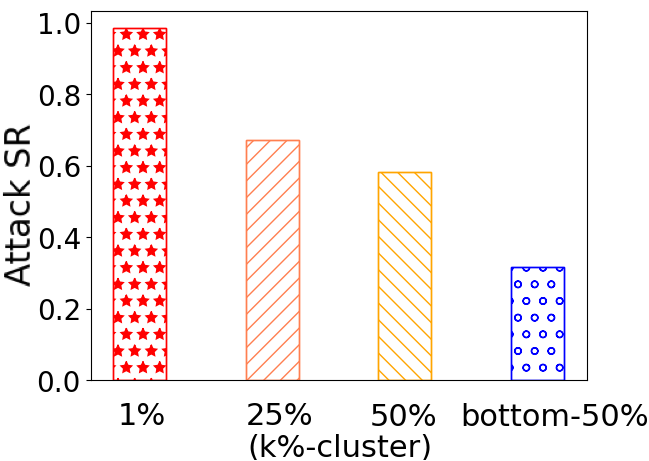}}\hspace{0cm}
\subfigure[Sentiment]{\includegraphics[scale=0.16]{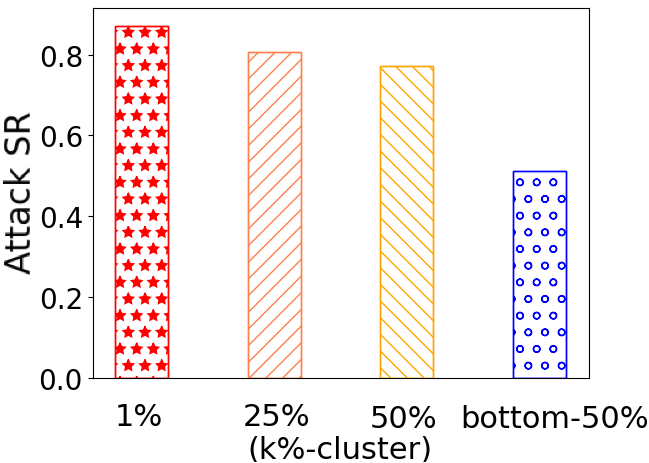}} 
      \caption{Label distributions and Attack SR.} 
      \label{fig:cosine}
\end{figure}

To answer the question, we examine the proximity between $X$ and groups of benign clients at different backdoor risk levels.
These sets include the $1\%$-cluster, $25\%$-cluster, $50\%$-cluster, and the remaining bottom-$50\%$-cluster of benign clients. The $k\%$-cluster consists of all benign clients having top-$k\%$ scores (Eq.~\ref{score}) while excluding clients in all preceding clusters. For instance,  $50\%$-cluster includes top-50\% infected clients, but it excludes clients in 25\%-cluster and 1\%-cluster. Fig.~\ref{client-level-mainbody} shows the distribution of these infected client groups. We compute the average cosine similarity of their cumulative label distributions to $X$ to examine proximity, as follows: 

\vspace{-7.5pt}
\begin{small}
\begin{equation}
 CS_{k} = \frac{1}{|N_{k}|} \sum_{i \in k} Cos\big(\mathcal{P}_{CL}(D_i), \mathcal{P}_{CL}(D_a) \big)
\end{equation}
\end{small}
\vspace{-5pt}

\noindent where $k$ is the $k$\%-cluster   of infected clients, $|N_{k}|$ is the number of clients in $k$, and $Cos()$ is cosine similarity function.  $\mathcal{P}_{CL} (D_i)$ and $\mathcal{P}_{CL} (D_a)$ are the local data's cumulative label distributions of client $i$ and the auxiliary data $D_a$ used to train 
$X$, 
where $\mathcal{P}_{CL} (\cdot) = [N_j]_{j \in [1, L]}$ and $N_j$ is the sum of numbers of data samples with labels from $1$ to $j$ (i.e., $N_j = \sum_{q = 1}^j N_q$). 

\begin{figure}[t]
\captionsetup[subfigure]{labelformat=empty}
  \centering
\subfigure[]{\includegraphics[scale=0.2]{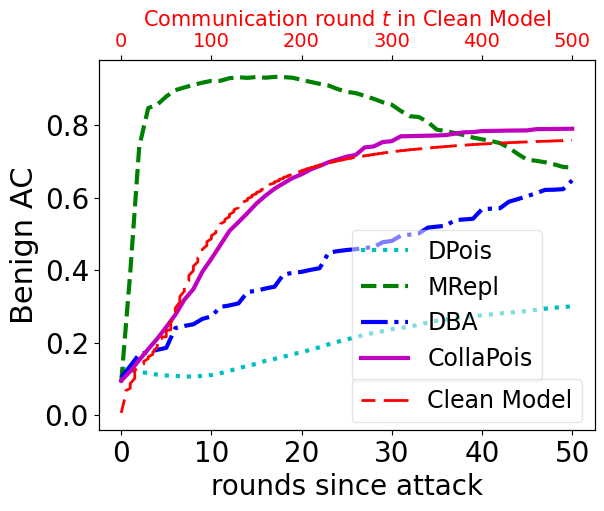}}\hspace{0.1cm}  
\subfigure[]{\includegraphics[scale=0.2]{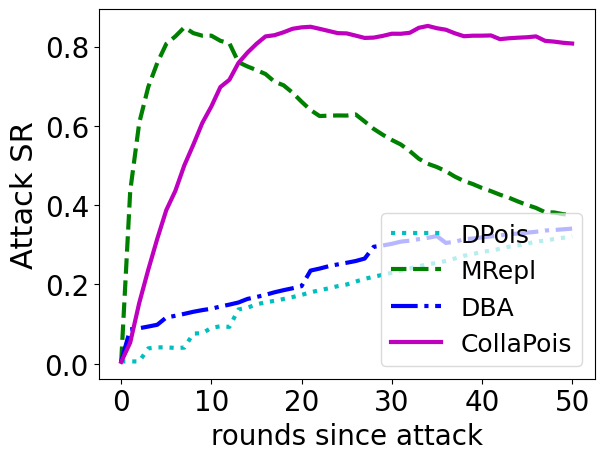}} \vspace{-10pt} 
      \caption{Benign AC and Attack SR as a function of training rounds. ($1\%$ compromised clients, $\alpha = 0.01$, FEMNIST)}
      \label{fig:training rounds}
\end{figure}\setlength{\textfloatsep}{2pt}

Fig.~\ref{fig:cosine} shows that benign clients with local data more aligned with the compromised clients' auxiliary data $D_a$ (higher cosine similarity) are more vulnerable (higher Attack SR). This is because their gradients align more closely with malicious gradients, making them more likely to be influenced by   $X$ and highly susceptible to the backdoor attack.


In the FEMNIST dataset, the 1\%-cluster infected clients have the highest $CS_1$  ($0.95$) and average Attack SR ($98.49\%$) compared with $CS_{25} =0.91$ and $CS_{50}=0.90$ and $67.11\%$ and $58.21\%$ Attack SR of the 25\%-cluster and 50\%-cluster infected clients, respectively. The bottom-50\%-cluster infected clients have the lowest $CS_{\text{bottom-} 50}$ ($0.85$), and consequently the lowest Attack SR (31.60\%). The Sentiment dataset exhibits a similar trend. However, the gap between the 1\%-cluster infected clients with the 25\%--cluster and the 50\%-cluster is smaller. This is because the clustering of infected clients near auxiliary data $D_a$ results in lower variability in cosine similarity and Attack SR across client groups (Fig.~\ref{fig:cosine}b). Similar trends are observed across  datasets and FL mechanisms.

\textbf{Stealthy and Longevity Attack.} Fig.~\ref{fig:training rounds} shows Attack SR and Benign AC over training rounds.  Unlike \textsc{MRepl} with sudden shifts (e.g., Benign AC raises from 39.21\% to 74.11\% in one round), \textsc{CollaPois} maintains consistently higher and long-lasting Attack SR across many rounds, with only a negligible 1\% drop after 40 rounds compared to \textsc{MRepl}'s 40\% decline.
Importantly, \textsc{CollaPois} converges significantly faster than \textsc{DPois} and \textsc{DBA}. This aligns with its two key properties: (1) it pulls FL models consistently toward  $X$, enhancing effectiveness, and (2) once near the Trojan-infected area, models are hard to reverse. This highlights \textsc{CollaPois} as  highly effective and stealthy compared to baseline attacks.



\vspace{-5pt}
\section{Discussion} 
\vspace{-2.5pt}

This section aims to contribute our insights into the issues at hand and potentially guide future research endeavors.

\textbf{Attack Perspective}. 
\textsc{CollaPois} poisons specific clients using divergent data. To escalate this threat, we target high-value clients only, minimizing detection. A ``semi-ready''  Trojaned model $X$ activates after updates from these clients, using (1) auxiliary data to approximate client behavior or (2) aggregated updates over multiple rounds  to detect client-specific patterns, boosting both attack precision and stealth.


\textbf{Defense Viewpoint}. Evaluation shows current defenses in (personalized) FL are largely ineffective against highly divergent client data. Methods like DP and NormBound lack protection, while Krum and RLR harm  model utility. Effective defenses for these challenges remain largely unexplored.

Our study shows that while benign client updates tend to cancel out, compromised clients can coordinate to propagate backdoors. Existing defenses like DP and NormBound fail to manage divergence, whereas Krum and RLR overly constrain it, harming benign performance. More balanced model update strategies offer promising alternatives.


\setlength{\textfloatsep}{5pt}

\section{Conclusion}

This study proposes a novel backdoor attack called \textsc{CollaPois} that exploits diverse data distribution among clients in Federated Learning (FL). Through theoretical analysis and extensive empirical experiments, we demonstrate the effectiveness, practicality, and stealthiness of \textsc{CollaPois}. We show that even with a small number of compromised clients, \textsc{CollaPois} can successfully converge the FL model around a pre-trained Trojaned model. It achieves higher backdoor attack success rates when clients exhibit greater data diversity, and it impairs the server's ability to detect suspicious behaviors. Furthermore, \textsc{CollaPois} can bypass current backdoor defenses, particularly when clients possess diverse data distributions. The evaluation results highlight that a mere 0.5\% of compromised clients can open a backdoor on 15\% of benign clients with an impressive success rate exceeding 70\% using state-of-the-art robust FL algorithms on benchmark datasets.

\section*{Acknowledgments}
This research was partially supported by the National Science Foundation (NSF) CNS-1935928, and the Qatar National Research Fund (QNRF) ARG01-0531-230438.



\bibliography{sample-base.bib}
\bibliographystyle{IEEEtran}

\newpage
\clearpage
\appendix

\subsection{Robust Federated Training Summary Table}
\label{app:compare defenses}

Table \ref{tab:compare defenses} summarizes  various
robust FL approaches.

\begin{table*}[t]
\footnotesize
\caption{Robust federated training algorithms against backdoor attacks.}  \vspace{-5pt}
\label{tab:compare defenses}
\centering
\begin{tabular}{|l|l|l|}
\hline
Approach & Method            & Description   \\
 \hline 
 \hline
\multirow{ 4}{*}{}& \makecell[l]{Krum / Multi-Krum \cite{blanchard2017machine} } & \makecell[l]{Score each update based on its closeness to its neighbors; \\Take the average of top N updates as aggregated update}  \\
\cline{2-3}
& Median GD \cite{yin2018byzantine} & Use the element-wise median as aggregated update \\
\cline{2-3}
\makecell[c]{Robust Aggregation } & Trim Mean GD \cite{yin2018byzantine} & \makecell[l]{Remove the top and bottom $\beta$ percentage of collected updates;
\\Use the element-wise mean as aggregated update} \\
\cline{2-3}
& SignSGD \cite{bernstein2018signsgd} & Adjust the server's learning rate based on the agreement of client updates \\ \cline{2-3}
& \makecell[l]{Robust Learning Rate \cite{ozdayi2021defending}}  & \makecell[l]{Count the updates in the same direction of  aggregated update for each element; \\
Flip the update in elements where the count  is smaller than the threshold} \\
\cline{2-3}
& Ditto \cite{li2021ditto} & Fine-tune the potentially corrupt global model on each client's private data \\ 
\hline
&  Norm Bound \cite{sun2019can} & \makecell[l]{Clip the gradients based on magnitude; Add Gaussian noise to the gradients} \\
\cline{2-3}
\makecell[c]{Model Smoothness} &  CRFL \cite{xie2021crfl}  & \makecell[l]{
 Clip model parameters to control model smoothness; Generate sample robustness certification} \\ 
 \cline{2-3}
 & FLARE \cite{wang2022flare} &  \makecell[l]{Estimate a trust score for each model update based on the differences between \\ all pair of updates; Aggregate model updates weighted by the trust scores } \\
\hline
\makecell[c]{Differential Privacy} & DP-optimizer \cite{hong2020effectiveness} & \makecell[l]{Clip the gradients collected from clients; Add Gaussian noise to the clipped gradients}  \\ 
\cline{2-3}
& User-level DP \cite{mcmahan2017learning} & Add sufficient Gaussian noise to model  updates for providing user-level DP \\
\hline
  \end{tabular} \par 
\end{table*}

\subsection{Proof of Theorem \ref{Theorem-angles}}
\label{proof: Theorem-angles}

\begin{proof}
Given the diverse data distribution among clients, their
gradient updates vary in direction and magnitude. To ensure
the attack's  effectiveness, it is necessary for the aggregated model  updates at each iteration $t$ to align with the direction of the aggregated malicious gradient $\sum_{i \in \mathcal{C} } \psi_c g_{\Delta_c}$. To capture both the direction and magnitude, we project all
gradients onto the direction of the aggregated malicious gradient. This leads to the following condition: 

 \begin{small}
\begin{equation}
     \sum_{i \in \mathcal{C} }  \psi_c g_{\Delta_c} + \sum_{i \in N \setminus \mathcal{C} }   g_{\Delta_i}   \ge 0,
       \label{condition-g-refined}
\end{equation}
\end{small}
where $g_{\Delta_i}$  is the projection of the gradient $\Delta_i$ into the direction of the malicious aggregated gradient $\Delta_c$ and $\psi_c$ is the dynamic learning rate ($\psi_c \sim \mathcal{U}[a, b]$). 
    
In worst-case scenarios where the benign gradients are oriented in the opposite direction to the aggregated malicious
gradient, Eq.~\ref{condition-g-refined} can be reformulated as follows:
\begin{small}
 \begin{equation}
        (\sum_{c \in \mathcal{C}}  \psi_c \cdot A_c)\vec{i}  - \sum_{i \in N \setminus \mathcal{C} } [\cos(\beta_i) \cdot A_i^b \vec{i}]    \ge 0,
         \label{condition-g-i}
    \end{equation}
    \end{small}
    where $\vec{i}$ is a unit vector representing the direction, and $A_{c}$ and $A^{b}_{i}$ are the magnitudes of the gradients from compromised and benign clients, respectively. 
    To circumvent gradient exploration and prevent the server from tracking the gradients to identify suspicious behavior patterns, we upper-bound the magnitude of the gradients by $A$ (i.e., $\max A_c = \max A_i^b = A$).  Then, Eq.~\ref{condition-g-i} becomes:
    \begin{small}
    \begin{equation}
        (\sum_{c \in \mathcal{C}}  \psi_c )   - \sum_{i \in N \setminus \mathcal{C} } [\cos(\beta_i)   ] \ge 0.
        \label{eq:sum-condition}
    \end{equation}
    \end{small}

To calculate $\sum_{i \in N \setminus \mathcal{C} } [\cos(\beta_i)   ] $,  by applying Maclaurin's theorem   to the cosine function (as in Trigonometry), we have: 
$\cos (\beta_i) = 1 - \frac{\beta_i^2}{2!}   + \frac{\beta_i^4}{4!} = \sum_{k=0}^{\infty} (-1)^{k} \frac{(\beta_i)^{2k}}{(2k)!}$. Therefore, we can approximate the term 
$\sum_{i \in N \setminus \mathcal{C} } \cos(\beta_i) $ with an error bounded by $\mathcal{O}(\frac{\sum_{i \in N\backslash\mathcal{C}} (\beta_{i})^4}{4!})$:
\begin{align}
\sum_{i \in N\backslash\mathcal{C}} \cos (\beta_{i} )\approx (|N| - |\mathcal{C}|) - \frac{\sum_{i \in N\backslash\mathcal{C}} (\beta_{i})^{2}}{2}.
\label{eq:cond-beta}
\end{align}
Then, Eq.~\ref{eq:sum-condition} becomes:
\begin{small}
 \begin{equation}
        (\sum_{c \in \mathcal{C}}  \psi_c )   - \Big( (|N| - |\mathcal{C}|) - \frac{\sum_{i \in N\backslash\mathcal{C}} (\beta_{i})^{2}}{2} \Big) \ge 0.
        \label{eq:sum-condition2}
    \end{equation}
    \end{small}


Finding a closed-form solution for $|\mathcal{C}|$ to satisfy the condition in Eq.~\ref{eq:sum-condition2} is challenging, mainly because it is not feasible to precisely quantify the summations, as they are data-dependent. 
Therefore, we  \textit{(1)} approximate  $\sum_{c \in \mathcal{C}}  \psi_c $ with  $|\mathcal{C}| \cdot \frac{ (a+b)}{2}$ (since the mean of $  \psi_c \sim \mathcal{U}[a, b] $ is $\frac{a+b}{2}$) and \textit{(2)} replace $\sum_{i \in N\backslash\mathcal{C}} \beta_{i}^{2}$ with its  expectation, which is $\mathbb{E}(\sum_{i \in N\backslash\mathcal{C}} \beta_{i}^{2})$. As a result, we get, \vspace{-5pt}
\begin{scriptsize}
\begin{align}
 \nonumber & \mathbb{E } \Big( \sum_{i \in N\setminus \mathcal{C}} {\beta_i^2} \Big)  =\mathbb{E } \Big(\sigma^2  \frac{ \sum_{i \in N\setminus \mathcal{C}} {\beta_i^2} }{ \sigma^2}  \Big)    \\
 &\nonumber  = \sigma^2 \mathbb{E }  \Big[  \sum_{i \in N\setminus \mathcal{C}} \Big( \frac{(\beta_i - \mu_{\alpha})^2 + 2 \mu_{\alpha} \beta_i - \mu_{\alpha}^2}{ \sigma^2} \Big) \Big] \\
    \nonumber &=  \sigma^2 \mathbb{E } \Big( \sum_{i \in N\setminus \mathcal{C}} (\frac{\beta_i-\mu_{\alpha} }{ \sigma})^2 \Big)    + 2  \sigma^2  \mathbb{E } \Big( \frac{\mu_{\alpha}}{\sigma^2} \sum_{i \in N\setminus \mathcal{C}} \beta_i \Big) - \mathbb{E } \Big( (|N|-|\mathcal{C}|) \mu_{\alpha}^2  \Big)\\
  \nonumber   &  =\sigma^2 (|N|-|\mathcal{C} |) + 2\mu_{\alpha} (|N|-|\mathcal{C} |) \mu_{\alpha} - (|N|-|\mathcal{C} |) \mu_{\alpha}^2     = (|N|-|\mathcal{C} |) ( \sigma^2 + \mu_{\alpha}^2  ).
  \label{condition-beta}
\end{align}
\end{scriptsize}

From Eqs.~\ref{eq:sum-condition2} and  \ref{condition-beta}, we have:

\begin{small}
\begin{align}
   \nonumber   & |\mathcal{C}| \cdot \frac{ (a+b)}{2} - \Big( 
|N| - |\mathcal{C}| - \frac{(|N| - |\mathcal{C}|) ( \sigma^2 + \mu_{\alpha}^2  )}{2} \Big) \ge 0 \\
\Leftrightarrow & |\mathcal{C}| \ge \frac{2-\sigma^2 - \mu_{\alpha}^2}{ a+b +2-\sigma^2 - \mu_{\alpha}^2} |N|.
\end{align}
\end{small}

Therefore, Theorem \ref{Theorem-angles} holds.
\end{proof}


\subsection{Proof of Theorem \ref{ConvergenceBound}}
\label{proof: ConvergenceBound}

\begin{proof}
At the round $t'$, $\bigtriangleup \theta_c^{t'} = \psi_c^{t'}[X - \theta^{t'}]$. This is equivalent to $X = \frac{\bigtriangleup \theta_c^{t'}}{\psi_c^{t'}} + \theta^{t'}$. 
In round $t$, according to the findings of Theorem \ref{Theorem-angles}, the global model is expected to be a more severely poisoned model for the compromised client $c$: $\theta^t = \bigtriangleup \theta_c^{t'} + \theta^{t'} + \zeta$. To quantify the distance between the global model $\theta^t$ and the Trojaned model $X$, we subtract $X$ from $\theta^t$ as follows: $\theta^t - X = (1 - \frac{1}{\psi_c^{t'}})\bigtriangleup \theta_c^{t'} + \zeta$. Hence, we can bound the $l_2$-norm of the distance $\theta^t - X$ as follows:
\begin{small}
\begin{align}
\|\theta^t - X\|_2 & = \|(1 - \frac{1}{\psi_c^{t'}})\bigtriangleup \theta_c^{t'} + \zeta \|_2 \leq (\frac{1}{a} - 1) \| \bigtriangleup \theta_c^{t'} \|_2 + \|\zeta\|_2 
\end{align}
\end{small}
Consequently, Theorem \ref{ConvergenceBound} holds.
\end{proof}

\subsection{Proof of Theorem \ref{XapproximationError}}
\label{proof: XapproximationError}

\begin{proof} We have that $Error = \| X' -X\|_2 $
\begin{small}
\begin{align}
  \nonumber  = \| \sum_{c \in \bar{\mathcal{C}}} \frac{\theta^t_c}{p|\mathcal{C}|} + \sum_{i \in \bar{L} } \frac{\theta^t_i}{(1-p)(|N|-|\mathcal{C}|)} - X \|_2  = \| \sum_{c \in \bar{\mathcal{C}} \cup \bar{L} } \frac{\theta^t_c}{|\mathcal{C}|} - X \|_2,
\end{align}
\end{small}

\noindent  \text{in which } $ \| X' - X \|_2  \ge  \| \sum_{c \in \bar{\mathcal{C}}  } \theta_t^c /(p |\mathcal{C}|) - X \|_2 $

\begin{align}
\label{lowerbound-cond}
 \quad  \quad \quad  \quad  \quad   \quad    =  \| \sum_{c \in \bar{\mathcal{C}} } \frac{\bigtriangleup \theta_c^t }{p |\mathcal{C}| \psi^c_t } \|_2 
\ge  \| \sum_{c \in \bar{\mathcal{C}} } \frac{\bigtriangleup \theta_c^t }{p|\mathcal{C}| b } \|_2,
\end{align}

{\small
  \begin{equation}  
    \text{and } \| \sum_{c \in \bar{\mathcal{C}} \cup \bar{L} } \theta^t_c/|\mathcal{C}| - X \|_2 
    \le \arg\max_{L \subseteq N  \text{ s.t. } |L|=|\mathcal{C}| } \| \sum_{i \in L } \theta^t_i / |L| - X \|_2.
    \label{upperbound-cond}
\end{equation}
}

From Eqs.~\ref{lowerbound-cond} and \ref{upperbound-cond}, we have the following error bounds:
\begin{equation}
 \big\| \sum_{c \in \bar{\mathcal{C}} } \frac{\bigtriangleup \theta^t_c}{p|\mathcal{C}| b } \big\|_2  \le  Error \le \arg\max_{L \subseteq N \text{  s.t. } |L|=|\mathcal{C}| }  \big\| \sum_{i \in L } \frac{\theta^t_i}{|L|} - X \big\|_2.
\end{equation}
As a result, Theorem \ref{XapproximationError} holds.
\end{proof}

\subsection{Data and Model Configuration.} We conduct experiments on  Sentiment \cite{go2009twitter} and FEMNIST  \cite{caldas2018leaf} datasets. 
To control data distribution across clients in terms of classes and size of local training data, we leverage the symmetrical Dirichlet distribution with different values of the concentration parameter $\alpha \in [0.01, 100]$ as in \cite{gao2022feddc}. In short, the value of $\alpha$ is inversely proportional to the degree of diversity in data distribution. In Sentiment, we include $5,600$ clients with over $1$ million data samples. 
In FEMNIST, there are $3,400$ clients with $805,263$ samples. 
The client sampling rate $q=1\%$ and the dynamic learning rate $\psi \sim \mathcal{U}[0.9, 1]$. 
We designate class 0 as the targeted class for the attacker, denoted as $y^{Troj}$. We divide the data samples in each client into training (70\%), testing (15\%), and validation (15\%) sets. The combined validation set from all compromised clients serves as the auxiliary set for training the Trojaned model $X$.
In the following experiments, the attacker randomly compromised 0.1\%, 0.5\%, and $1\%$ of benign clients,  treating these small percentages of compromised clients as a practical threat   (\cite{drawingboard2022}). This percentage is below the required number of compromised clients, as indicated in Theorem \ref{Theorem-angles}. Our empirical experiments show that \textsc{CollaPois} maintains its effectiveness even when a smaller fraction of clients is compromised, thereby enhancing the overall efficacy, feasibility, and subtlety of our attack.

We adopt the model configuration described in \cite{shamsian2021personalized} for all the datasets. Specifically, we use a LeNet-based network  with two convolution and two fully connected layers for the local model, and a fully connected network with three hidden layers and multiple linear heads per target weight tensor. For the Sentiment dataset, we utilize the BERT model as the tokenizer and connect it with a two-layer fully connected network as the task head. We use SGD optimizer with the learning rate of $0.01$ for the aggregated global model and $0.001$ for the benign clients' local models.

\subsection{Visualization of WaNet \cite{nguyen2021wanet} Triggers }
\label{app:visualization wanet} \vspace{-5pt}

\begin{figure}[h]
      \centering
       \includegraphics[scale=0.2]{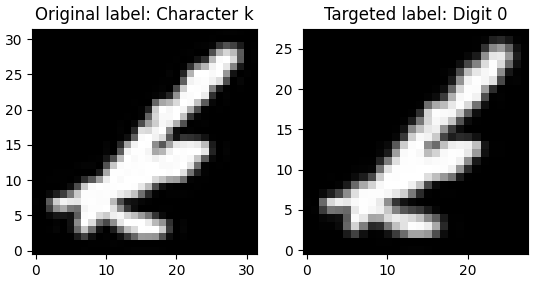}\vspace{-5pt}
      \caption{WaNet  \cite{nguyen2021wanet} in the FEMNIST dataset. Backdoor (right) and legitimate (left) samples  are almost identical.} \vspace{-10pt}
      \label{cifar10-vis}
\end{figure}




\subsection{Supplemental Results}\label{app:Supplemental Results}

\begin{figure}[t]
    \centering
   \subfigure[FedAvg-BenignAC]{\includegraphics[scale=0.275]{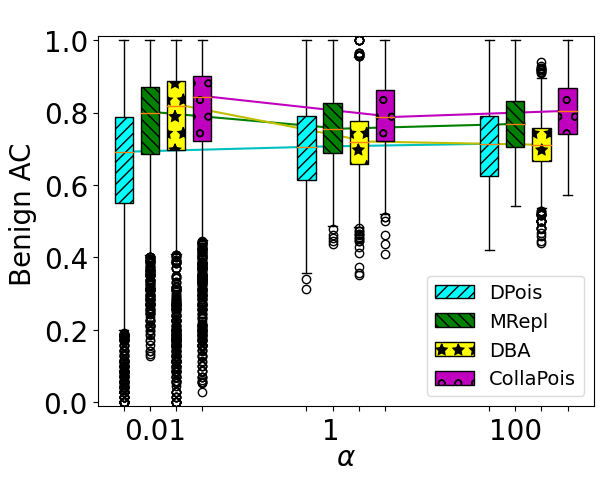}}%
      \vspace{-5pt}  \hfill
     \subfigure[FedAvg-AttackSR]{ \includegraphics[scale=0.275]{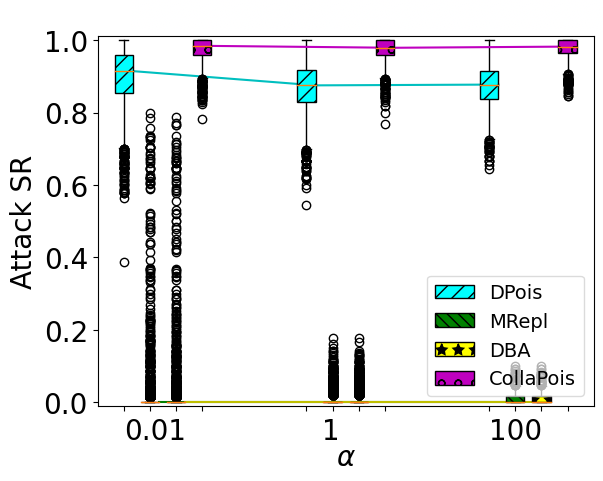} }\vspace{-5pt}
 \subfigure[FedDC-BenignAC]{\includegraphics[scale=0.265]{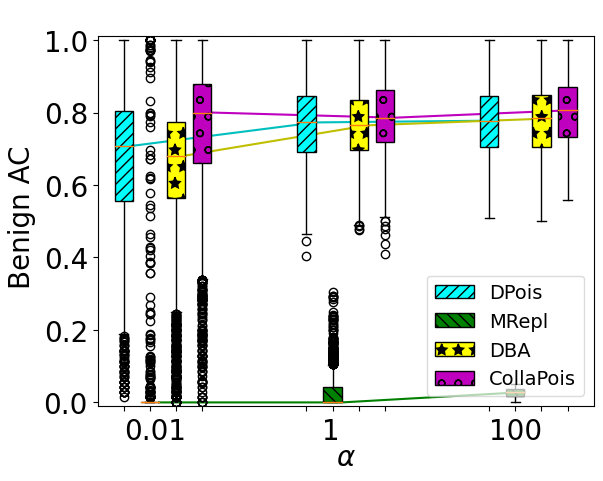}} \vspace{-5pt} \hfill
      \subfigure[FedDC-AttackSR]{\includegraphics[scale=0.265]{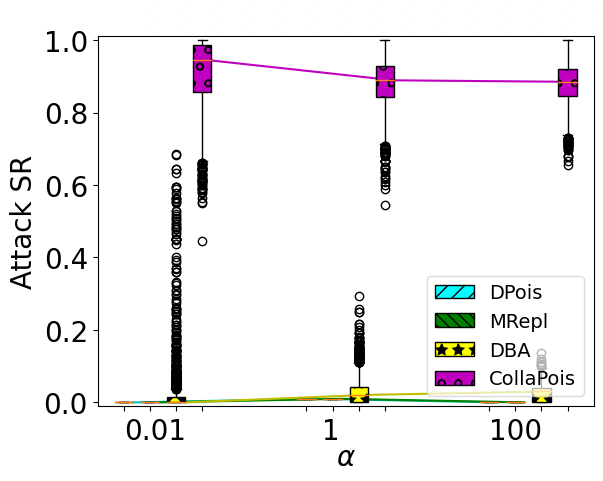}} \vspace{-5pt}
    \subfigure[MetaFed-BenignAC]{\includegraphics[scale=0.265]{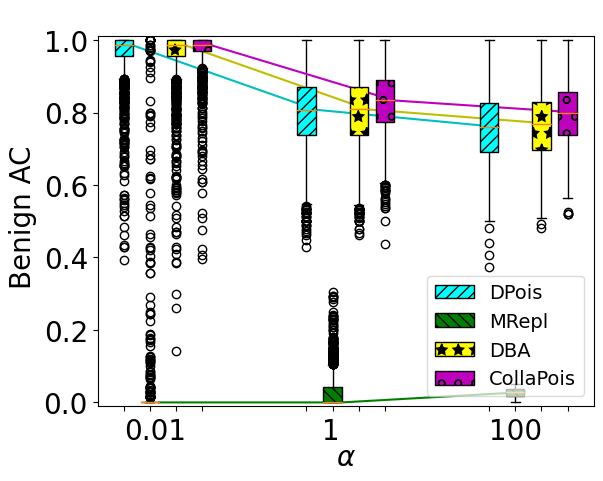} }\vspace{-5pt} \hfill
     \subfigure[MetaFed-AttackSR]{ \includegraphics[scale=0.265]{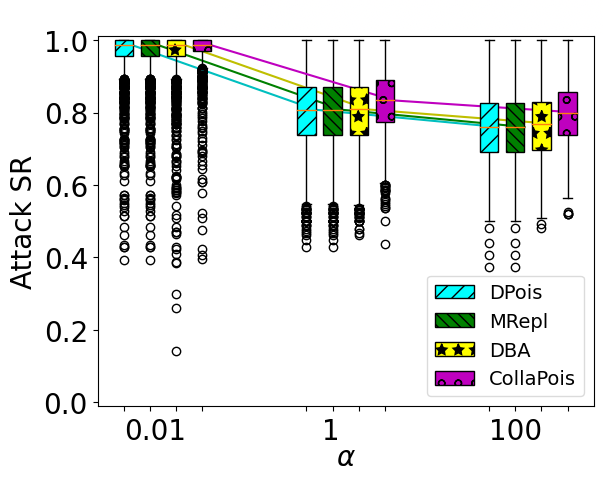} } 
    \caption{FedAvg, FedDC, and MetaFed under attacks ($1\%$ compromised clients) in the FEMNIST dataset.}
    \label{fig:all attacks-femnist-mainbody}  
\end{figure}\setlength{\textfloatsep}{2pt}

\begin{figure}[t]
    \centering
    \subfigure[FedAvg-BenignAC]{ \includegraphics[scale=0.27]{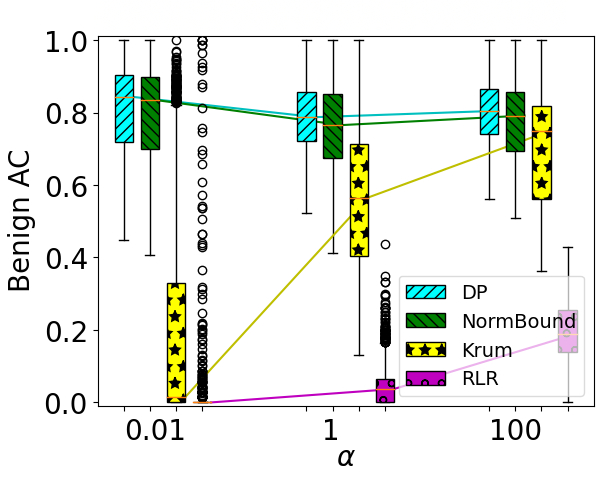} }%
       \vspace{-2.5pt}  \hfill
      \subfigure[FedAvg-AttackSR]{ \includegraphics[scale=0.27]{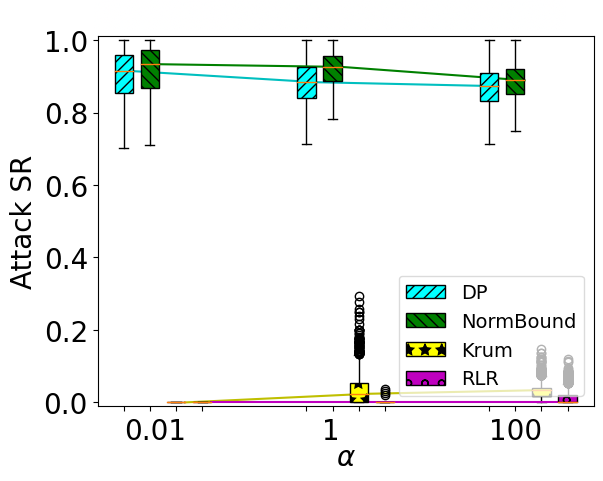}} \vspace{-5pt}
  \subfigure[FedDC-BenignAC]{
       \includegraphics[scale=0.27]{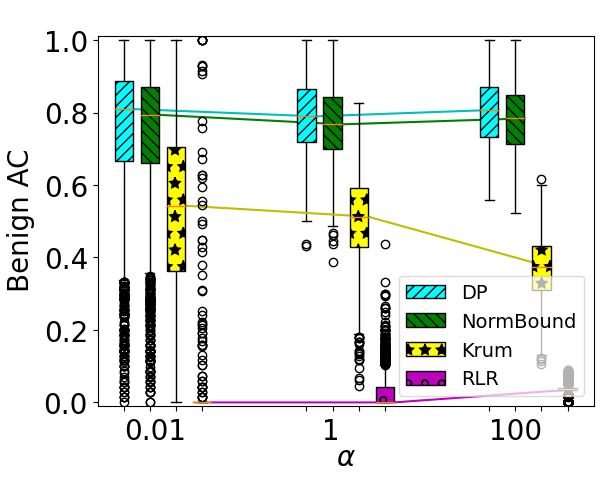}}\vspace{-5pt} \hfill
      \subfigure[FedDC-AttackSR]{ \includegraphics[scale=0.27]{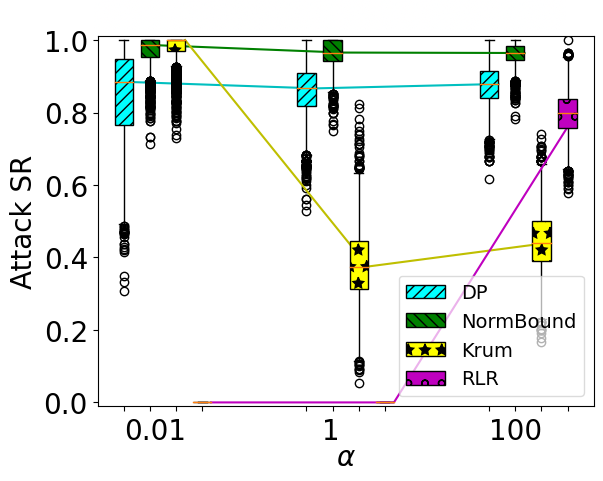}} \vspace{-5pt}
       \subfigure[MetaFed-BenignAC]{ \includegraphics[scale=0.27]{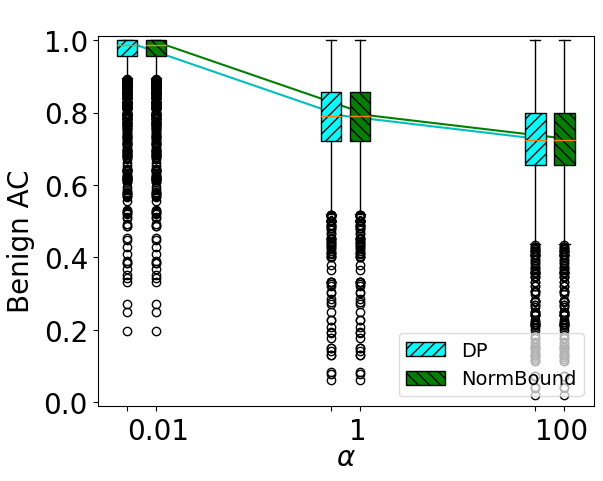} }\vspace{-2.5pt}\hfill
       \subfigure[MetaFed-AttackSR]{\includegraphics[scale=0.27]{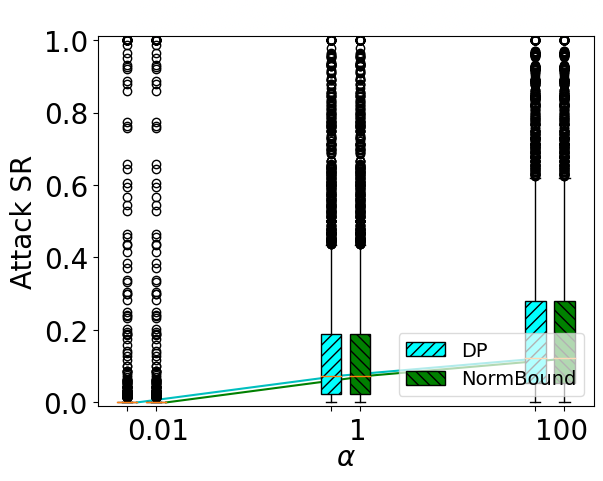} }
    \caption{\textsc{CollaPois} ($1\%$ compromised clients) under defenses for the FEMNIST dataset. (Krum and RLR are not applicable for MetaFed.)}
    \label{fig:all defenses-femnist-mainbody}  
\end{figure} \setlength{\textfloatsep}{2pt}



\begin{figure}[h]
    \centering
    \subfigure[FedAvg-BenignAC]{  \includegraphics[scale=0.27]{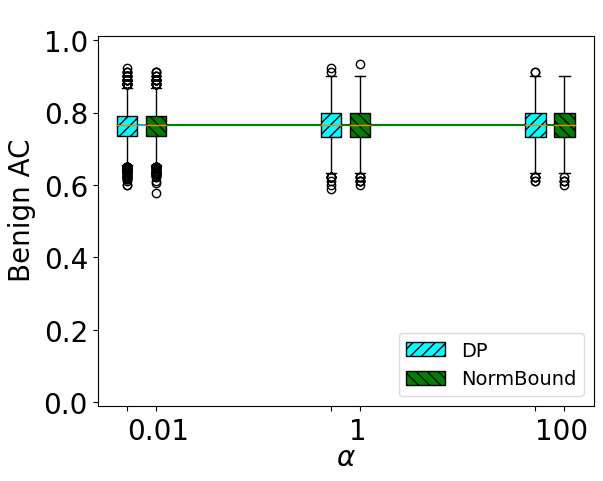}}%
        \hfill
     \subfigure[FedAvg-AttackSR]{  \includegraphics[scale=0.27]{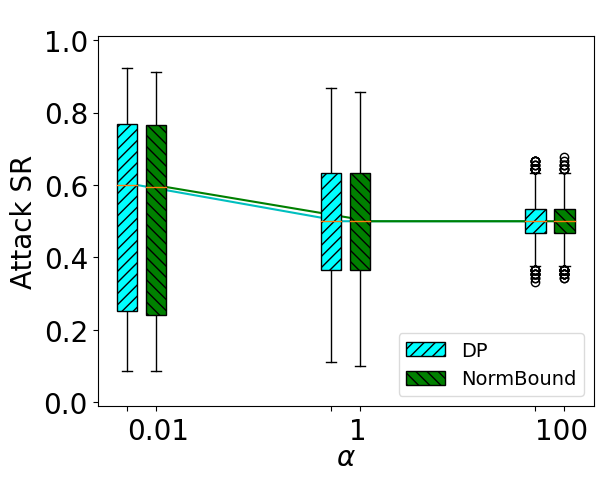}}
      \subfigure[FedDC-BenignAC]{  \includegraphics[scale=0.27]{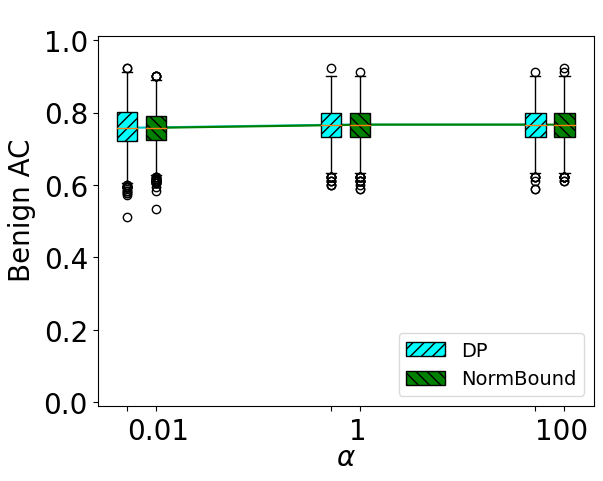}} \hfill
     \subfigure[FedDC-AttackSR]{  \includegraphics[scale=0.27]{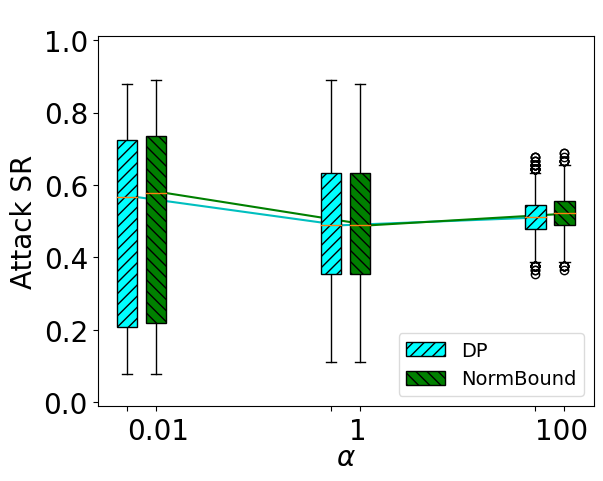}}
      \subfigure[MetaFed-BenignAC]{  \includegraphics[scale=0.27]{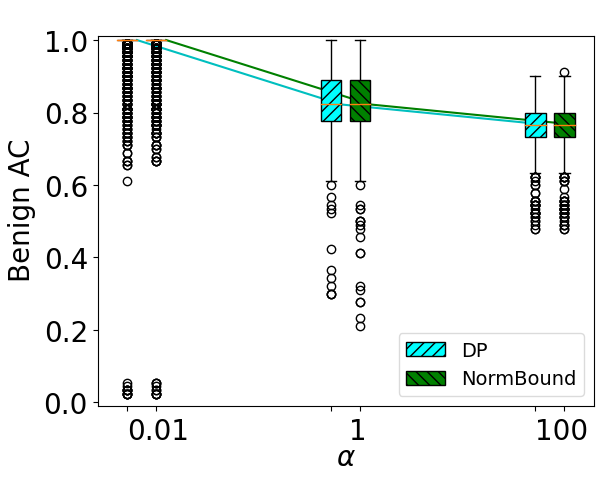} }\hfill
     \subfigure[MetaFed-AttackSR]{  \includegraphics[scale=0.27]{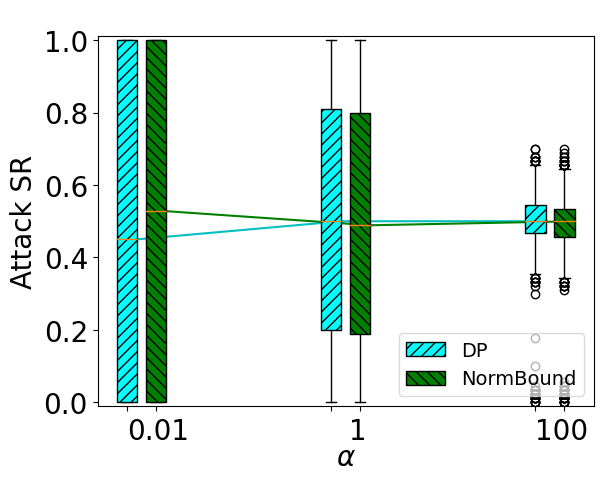}}
    \caption{\textsc{CollaPois} ($0.1\%$ compromised clients) under defenses for the Sentiment dataset in FedAvg, FedDC, and MetaFed.}
    \label{fig:all defenses-sent-app-0.1}  
\end{figure}

\begin{figure}[h]
    \centering
    \subfigure[FedAvg-BenignAC]{ 
      \includegraphics[scale=0.27]{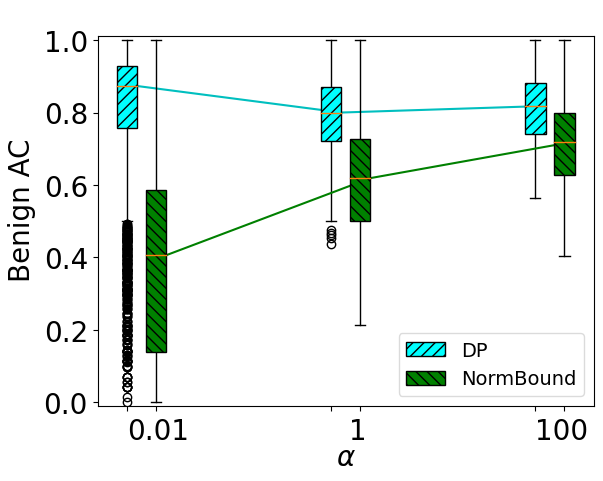}}%
        \hfill
     \subfigure[FedAvg-AttackSR]{  \includegraphics[scale=0.27]{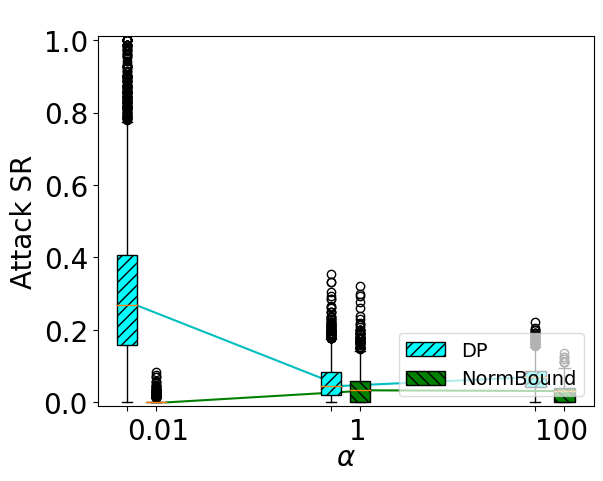}}
        \subfigure[FedDC-BenignAC]{ 
       \includegraphics[scale=0.27]{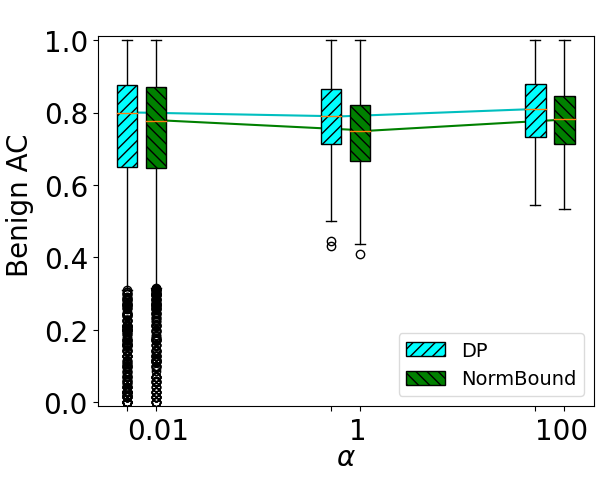}} \hfill
    \subfigure[FedDC-AttackSR]{   \includegraphics[scale=0.27]{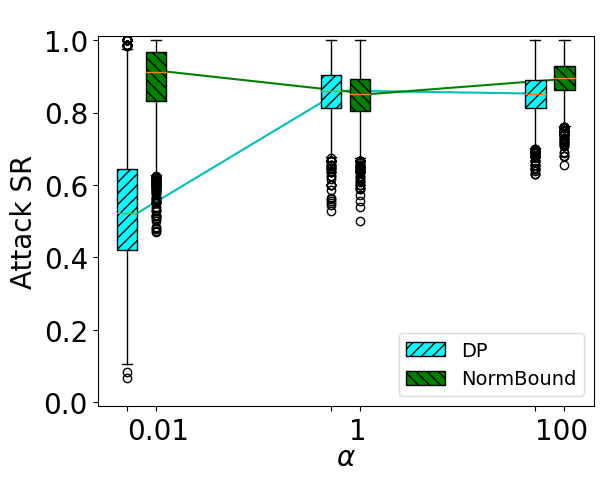}}
      \subfigure[MetaFed-BenignAC]{  \includegraphics[scale=0.27]{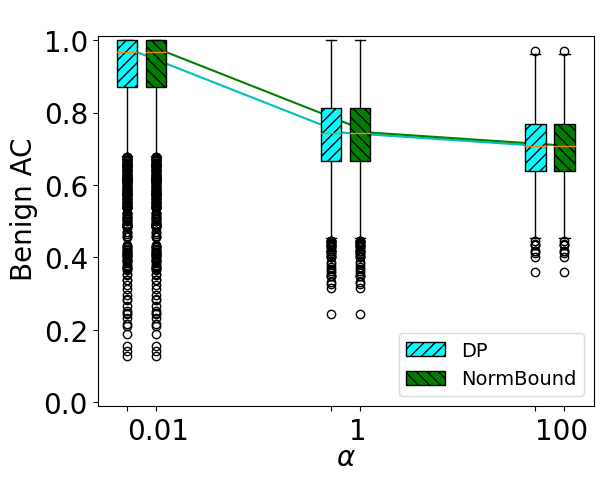} }\hfill
     \subfigure[MetaFed-AttackSR]{  \includegraphics[scale=0.27]{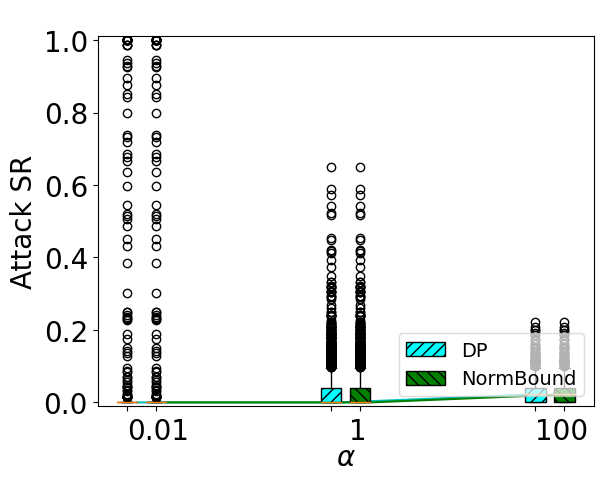}}
    \caption{\textsc{CollaPois} ($0.1\%$ compromised clients) under defenses for the FEMNIST dataset in FedAvg, FedDC, and MetaFed.}
    \label{fig:all defenses-femnist-app-0.1}  
\end{figure}



\begin{figure}[h]
    \centering
    \subfigure[FedAvg-BenignAC]{ \includegraphics[scale=0.27]{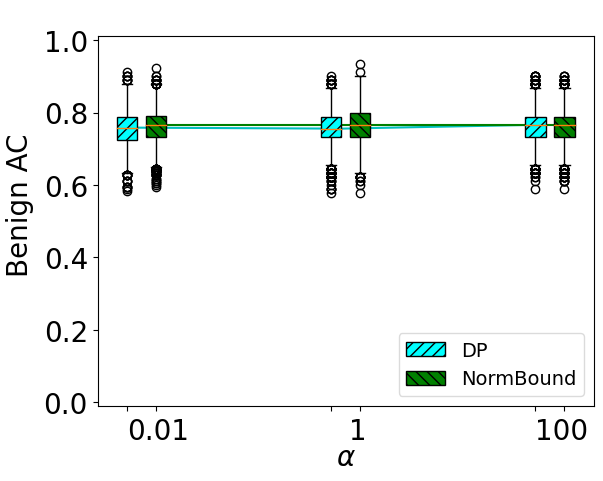}}%
        \hfill
     \subfigure[FedAvg-AttackSR]{ \includegraphics[scale=0.27]{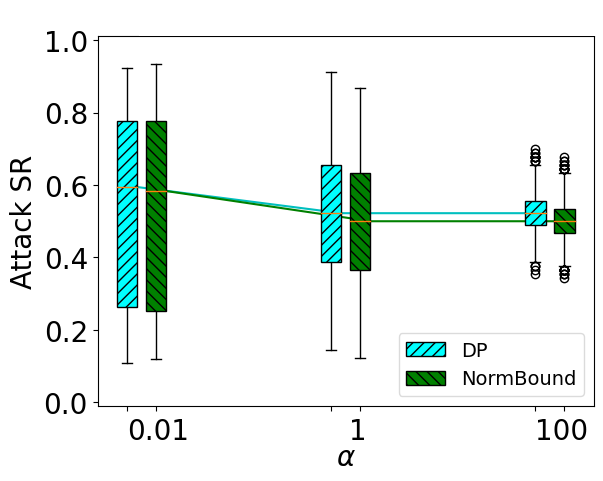}}
      \subfigure[FedDC-BenignAC]{ \includegraphics[scale=0.27]{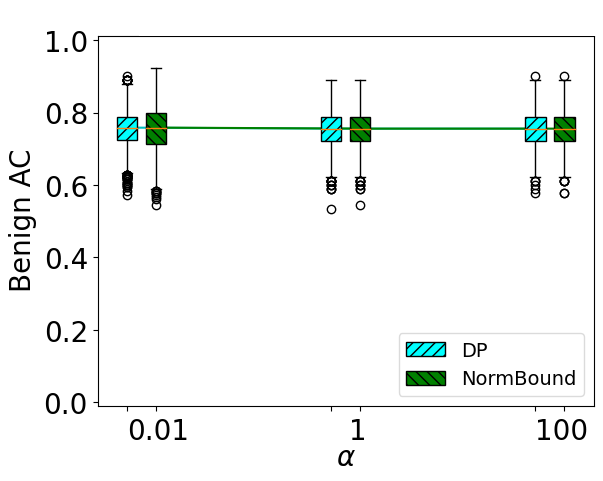} }\hfill
      \subfigure[FedDC-AttackSR]{\includegraphics[scale=0.27]{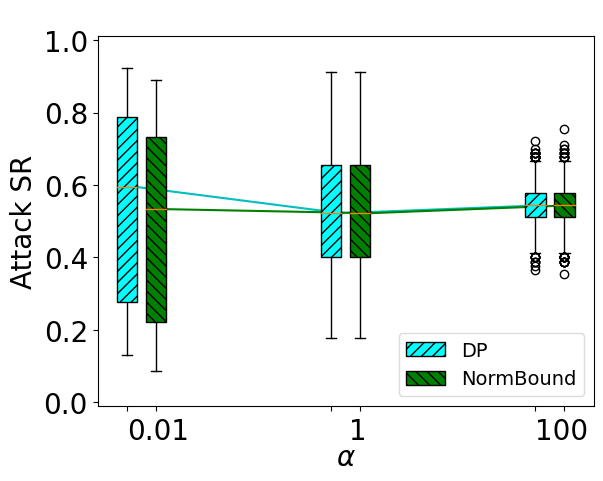}}
       \subfigure[MetaFed-BenignAC]{\includegraphics[scale=0.27]{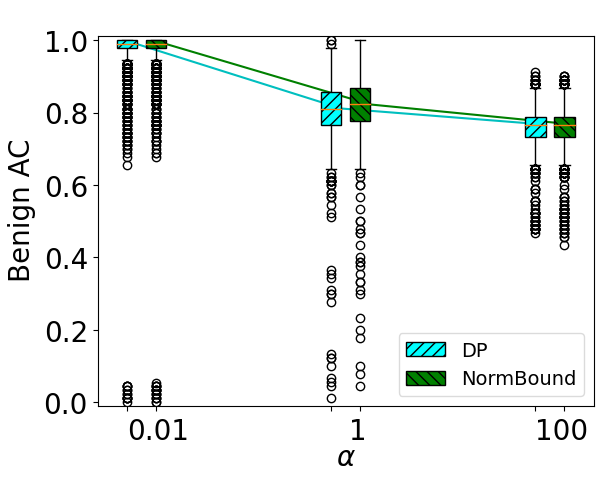}} \hfill
    \subfigure[MetaFed-AttackSR]{  \includegraphics[scale=0.27]{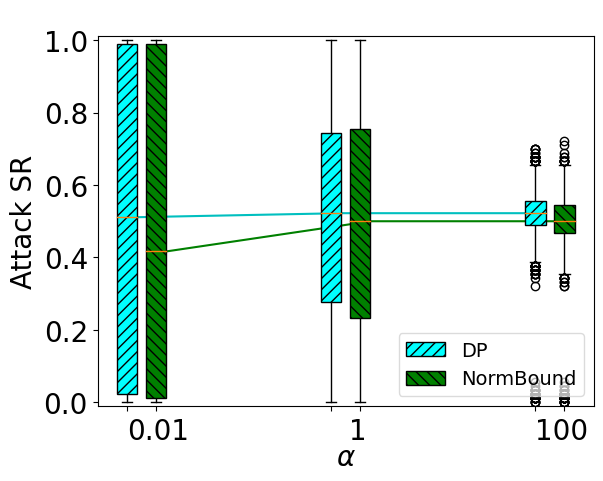}}
    \caption{\textsc{CollaPois} ($0.5\%$ compromised clients) under defenses for the Sentiment dataset in FedAvg, FedDC, and MetaFed.}
    \label{fig:all defenses-sent-app-0.5}  
\end{figure}

\begin{figure}[h]
    \centering
    \subfigure[FedAvg-BenignAC]{\includegraphics[scale=0.27]{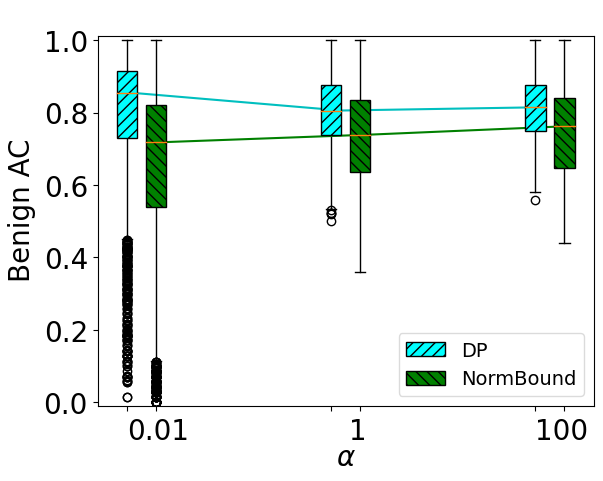}}%
        \hfill
      \subfigure[FedAvg-AttackSR]{\includegraphics[scale=0.27]{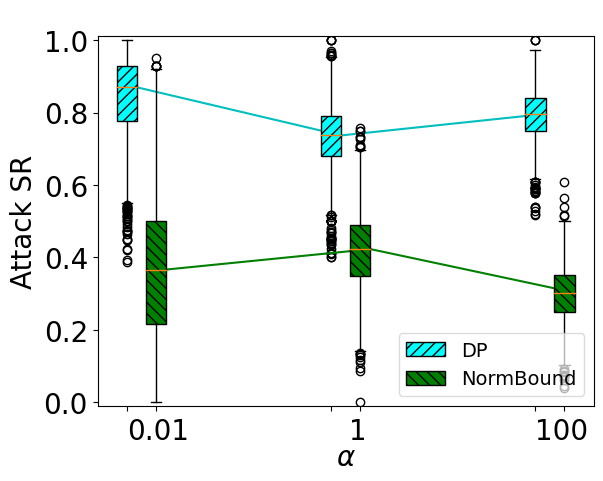}}
     \subfigure[FedDC-BenignAC]{  \includegraphics[scale=0.27]{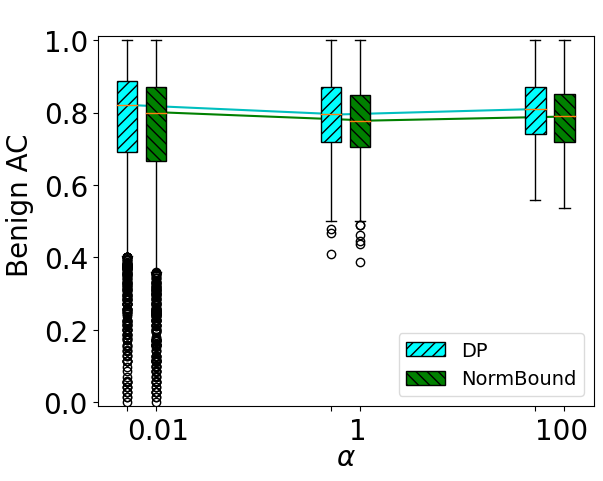} }\hfill
     \subfigure[FedDC-AttackSR]{\includegraphics[scale=0.27]{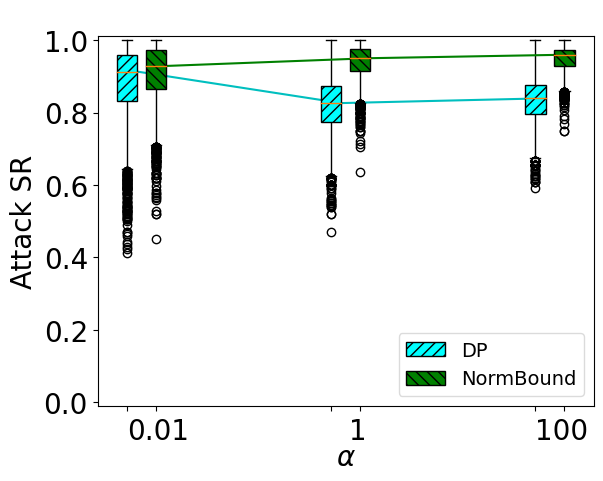}}
      \subfigure[MetaFed-BenignAC]{ \includegraphics[scale=0.27]{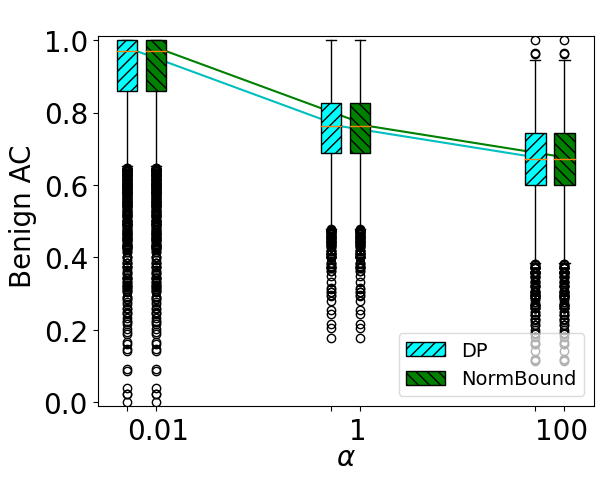} }\hfill
    \subfigure[MetaFed-AttackSR]{  \includegraphics[scale=0.27]{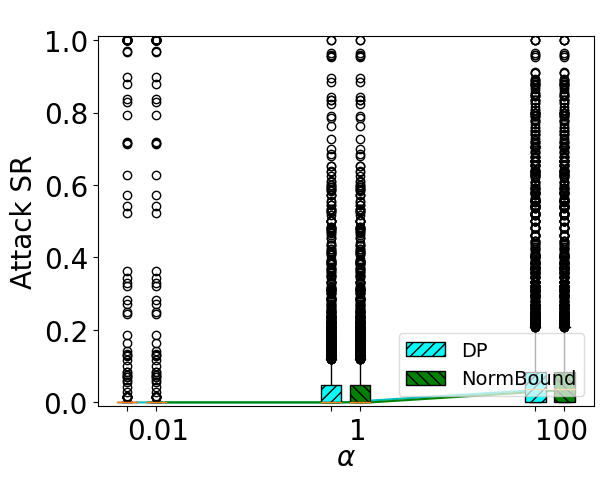}}
    \caption{\textsc{CollaPois} ($0.5\%$ compromised clients) under defenses for the FEMNIST dataset in FedAvg, FedDC, and MetaFed.}
    \label{fig:all defenses-femnist-app-0.5}  
\end{figure}

\begin{figure}[h]
    \centering
   \subfigure[FedAvg-BenignAC]{
      \includegraphics[scale=0.27]{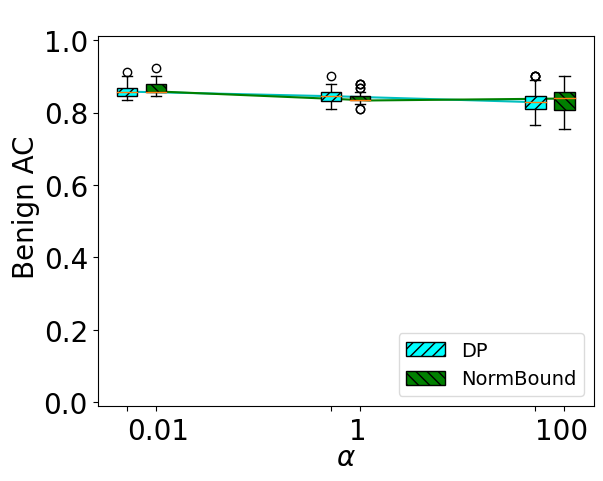}}%
        \hfill
     \subfigure[FedAvg-AttackSR]{ \includegraphics[scale=0.27]{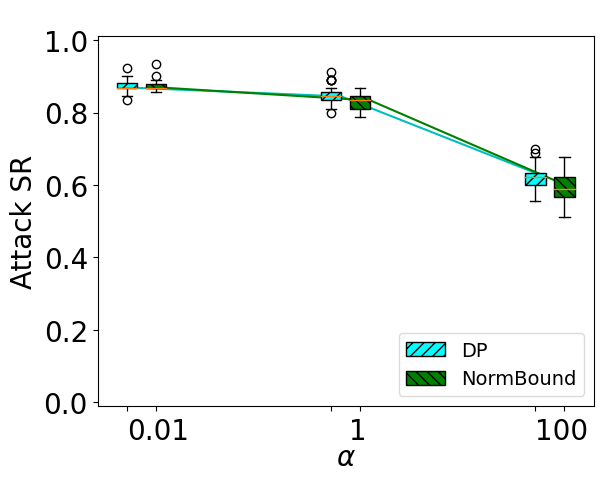}}
      \subfigure[FedDC-BenignAC]{ \includegraphics[scale=0.27]{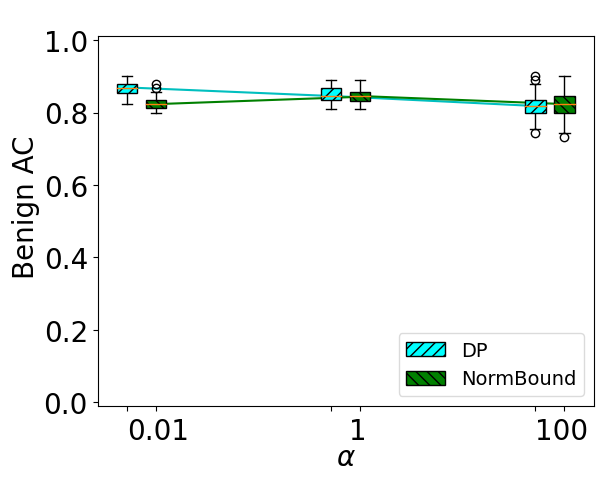} }\hfill
      \subfigure[FedDC-AttackSR]{\includegraphics[scale=0.27]{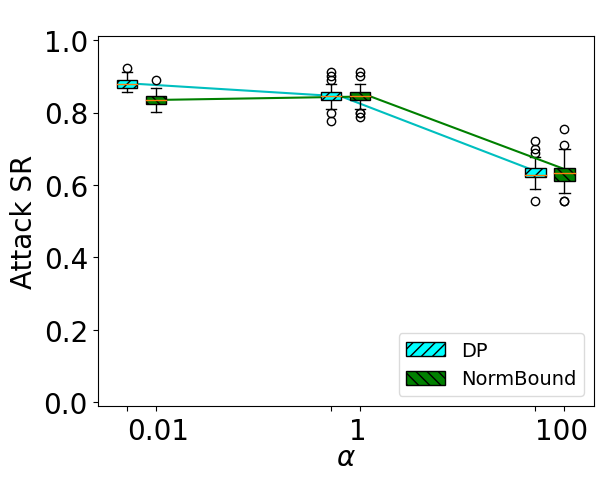}}
      \subfigure[MetaFed-BenignAC]{ \includegraphics[scale=0.27]{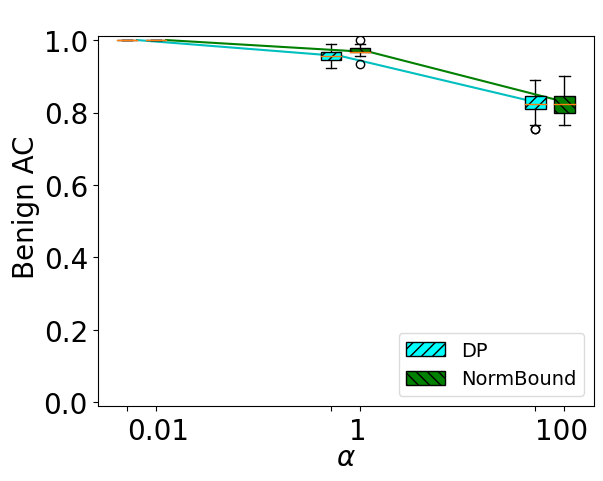}} \hfill
   \subfigure[MetaFed-AttackSR]{   \includegraphics[scale=0.27]{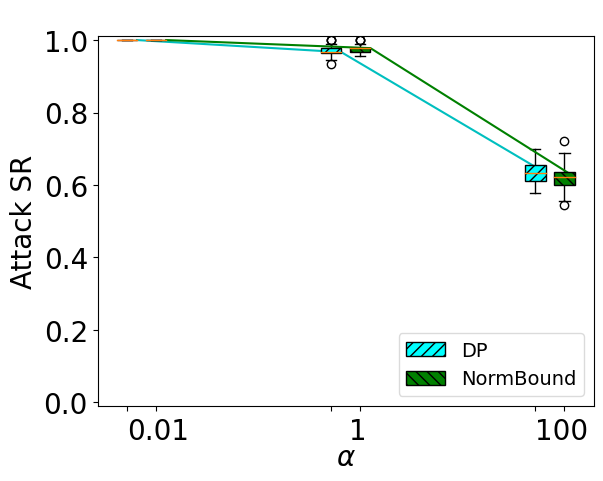}}
    \caption{\textsc{CollaPois} ($0.5\%$ compromised clients) under defenses for the Sentiment dataset in FedAvg, FedDC, and MetaFed. (Top-1\% clients)}
    \label{fig:all defenses-sent-app-0.5-top1}  
\end{figure}

\begin{figure}[h]
    \centering
    \subfigure[FedAvg-BenignAC]{ 
      \includegraphics[scale=0.27]{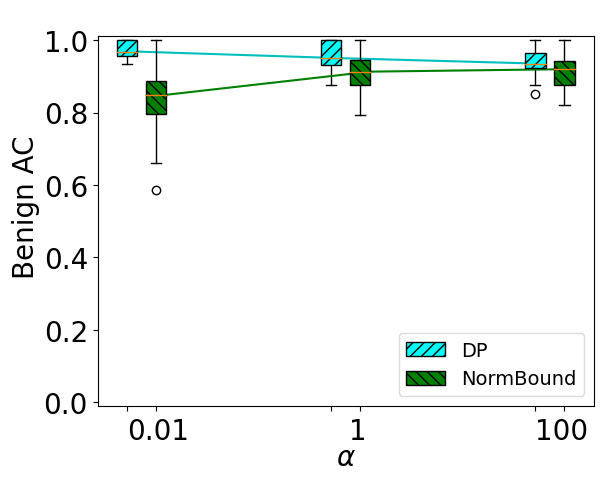}}%
        \hfill
      \subfigure[FedAvg-AttackSR]{ \includegraphics[scale=0.27]{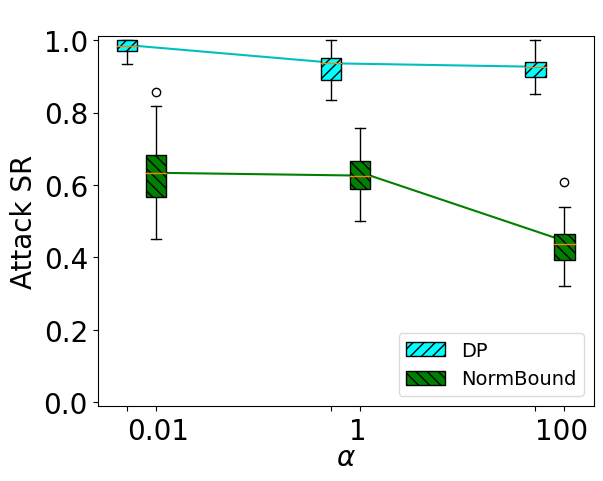}}
      \subfigure[FedDC-BenignAC]{  \includegraphics[scale=0.27]{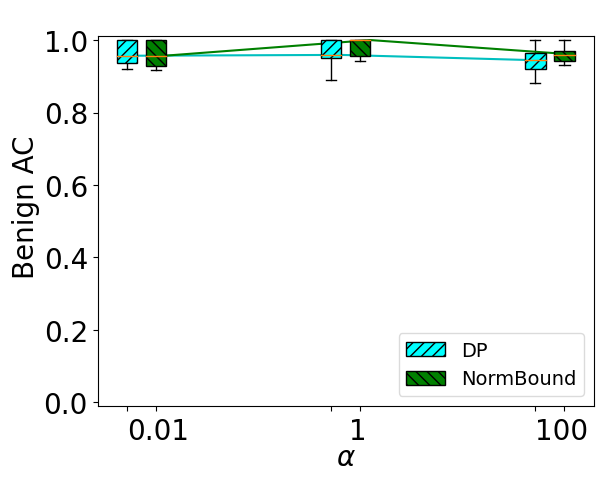} }\hfill
     \subfigure[FedDC-AttackSR]{  \includegraphics[scale=0.27]{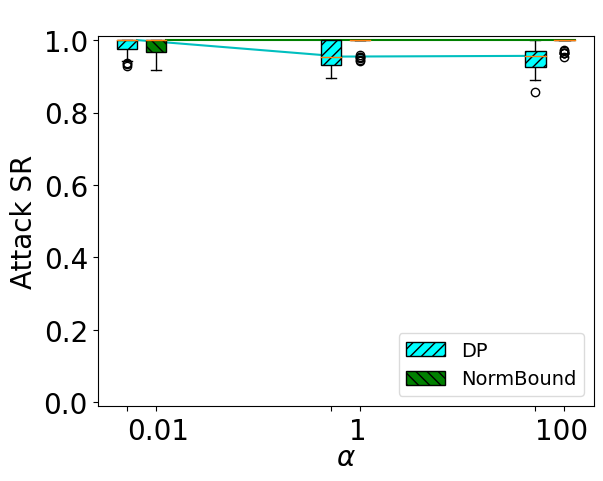}}
       \subfigure[MetaFed-BenignAC]{ \includegraphics[scale=0.27]{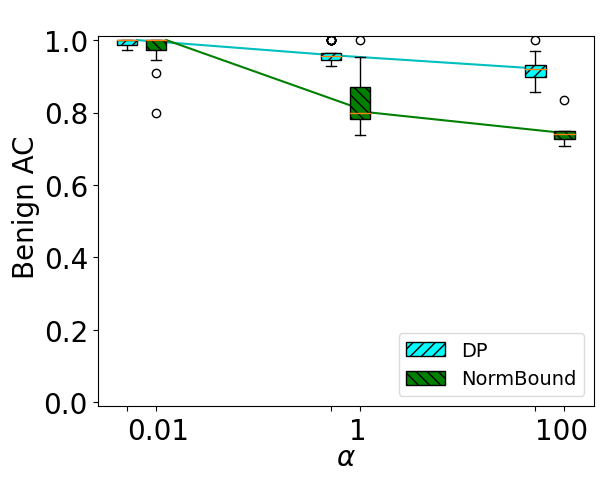} }\hfill
    \subfigure[MetaFed-AttackSR]{   \includegraphics[scale=0.27]{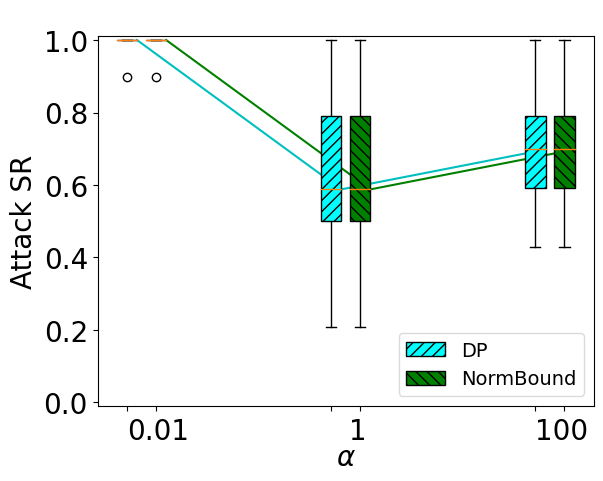} }
    \caption{\textsc{CollaPois} ($0.5\%$ compromised clients) under defenses for the FEMNIST dataset in FedAvg, FedDC, and MetaFed. (Top-1\% clients)}
    \label{fig:all defenses-femnist-app-0.5-top1}  
\end{figure}

\begin{figure}[h]
    \centering
      \subfigure[FedAvg-BenignAC]{  \includegraphics[scale=0.27]{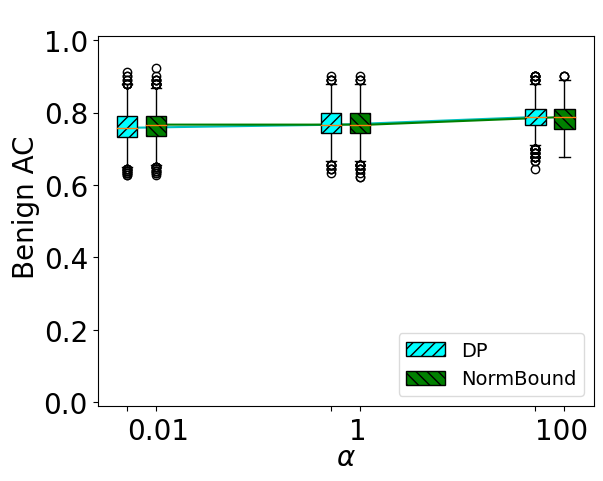}}%
        \hfill
       \subfigure[FedAvg-AttackSR]{ \includegraphics[scale=0.27]{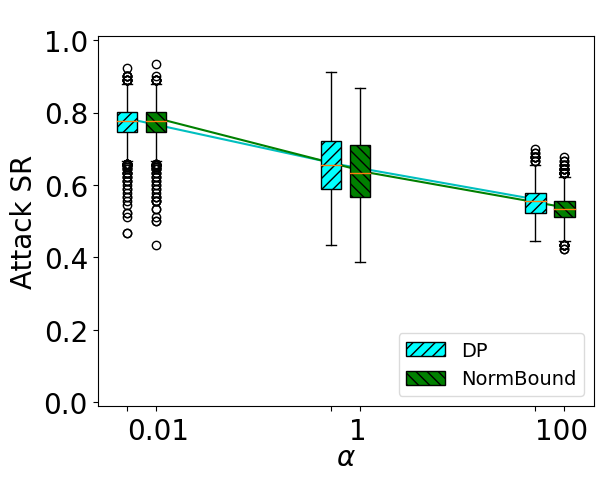}}
       \subfigure[FedDC-BenignAC]{  \includegraphics[scale=0.27]{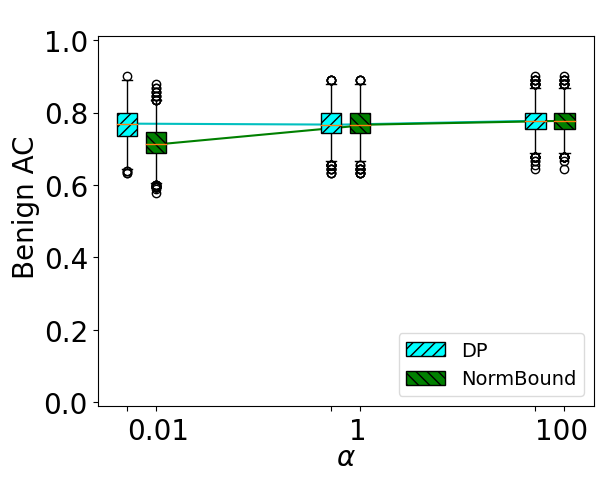}} \hfill
     \subfigure[FedDC-AttackSR]{   \includegraphics[scale=0.27]{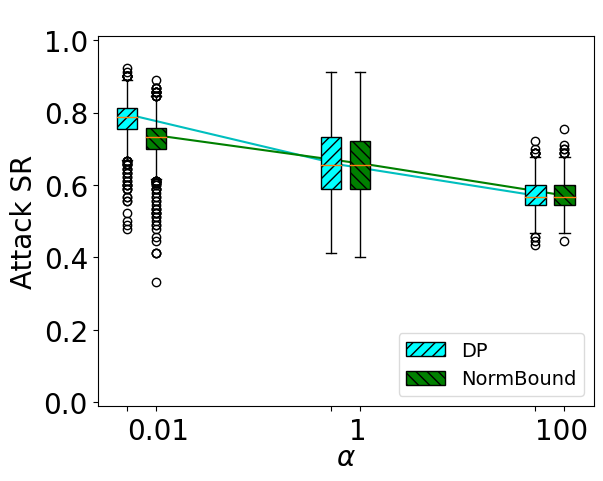}}
     \subfigure[MetaFed-BenignAC]{    \includegraphics[scale=0.27]{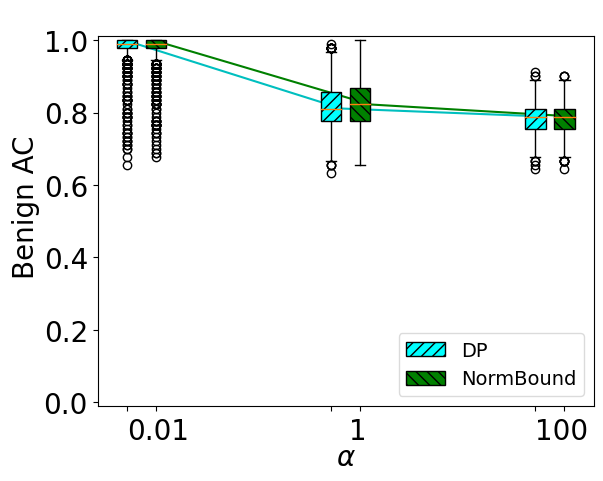}} \hfill
   \subfigure[MetaFed-AttackSR]{     \includegraphics[scale=0.27]{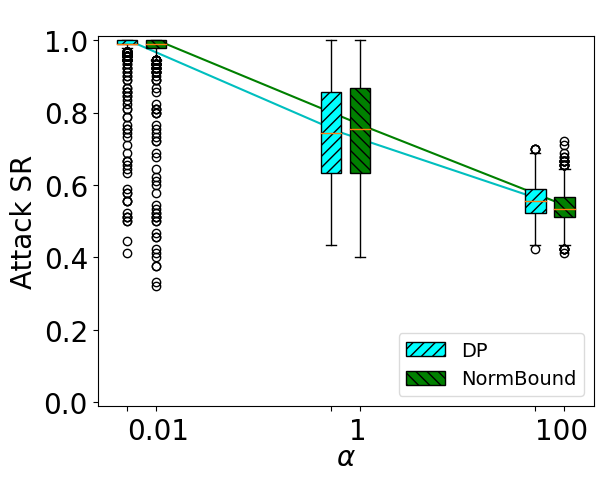}}
    \caption{\textsc{CollaPois} ($0.5\%$ compromised clients) under defenses for the Sentiment dataset in FedAvg, FedDC, and MetaFed. (Top-50\% clients)}
    \label{fig:all defenses-sent-app-0.5-top50}  
\end{figure}

\begin{figure}[h]
    \centering
      \subfigure[FedAvg-BenignAC]{ \includegraphics[scale=0.27]{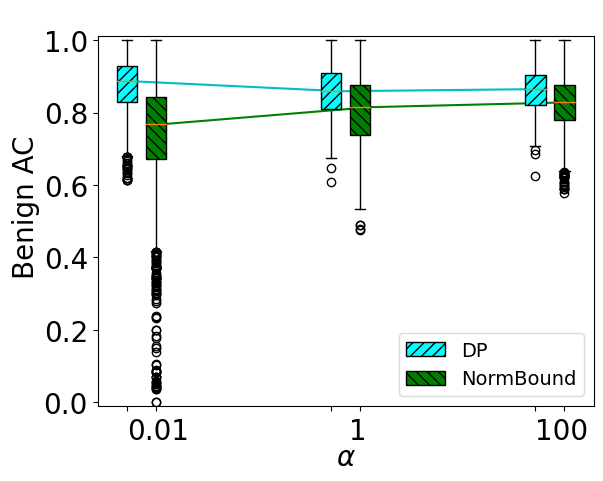} }%
        \hfill
      \subfigure[FedAvg-AttackSR]{ \includegraphics[scale=0.27]{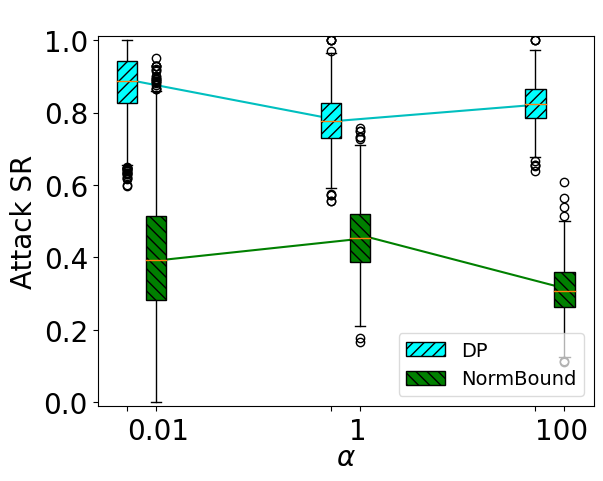}}
       \subfigure[FedDC-BenignAC]{ \includegraphics[scale=0.27]{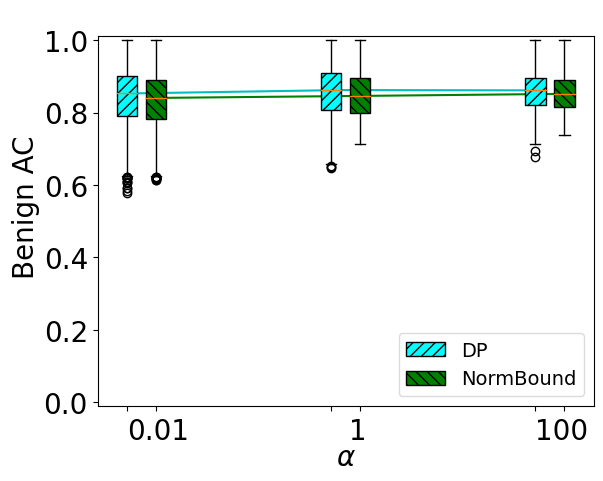}} \hfill
      \subfigure[FedDC-AttackSR]{ \includegraphics[scale=0.27]{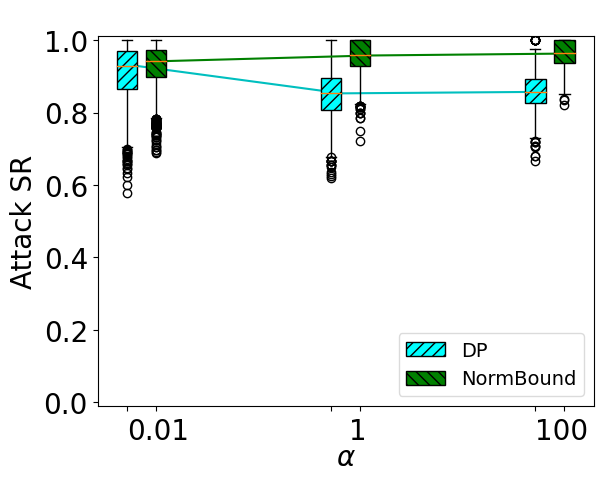}}
     \subfigure[MetaFed-BenignAC]{   \includegraphics[scale=0.27]{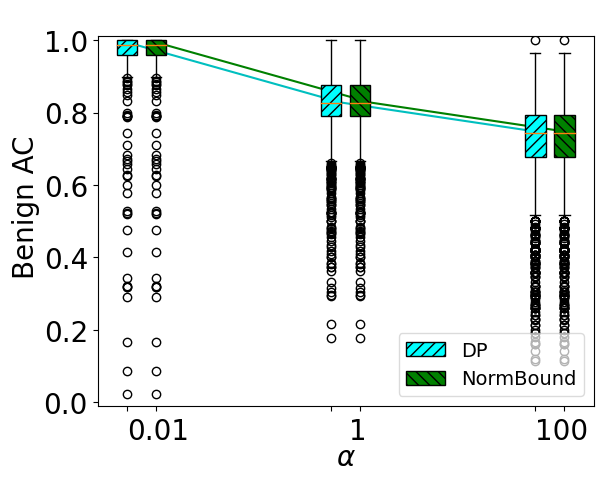} }\hfill
    \subfigure[MetaFed-AttackSR]{   \includegraphics[scale=0.27]{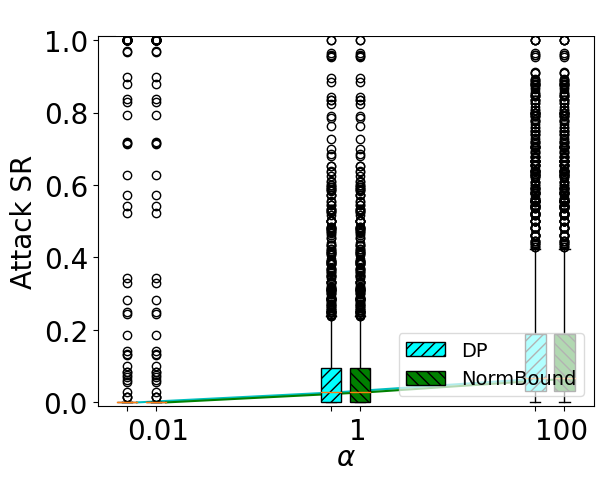}}
    \caption{\textsc{CollaPois} ($0.5\%$ compromised clients) under defenses for the FEMNIST dataset in FedAvg, FedDC, and MetaFed. (Top-50\% clients)}
    \label{fig:all defenses-femnist-app-0.5-top50}  
\end{figure}

\begin{figure}[t]
    \centering
   \subfigure[FedAvg-BenignAC]{     \includegraphics[scale=0.275]{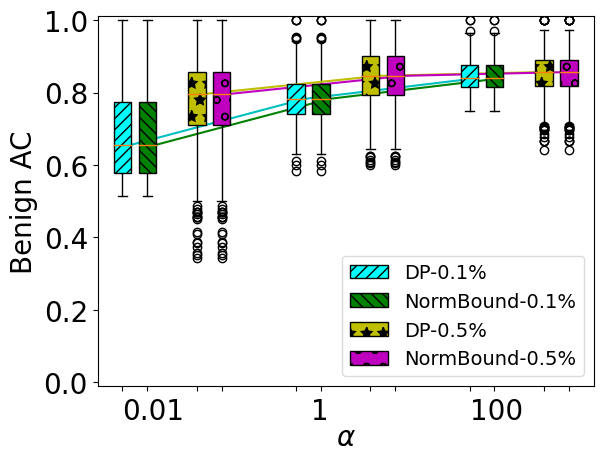}}%
        \hfill
    \subfigure[FedAvg-AttackSR]{    \includegraphics[scale=0.275]{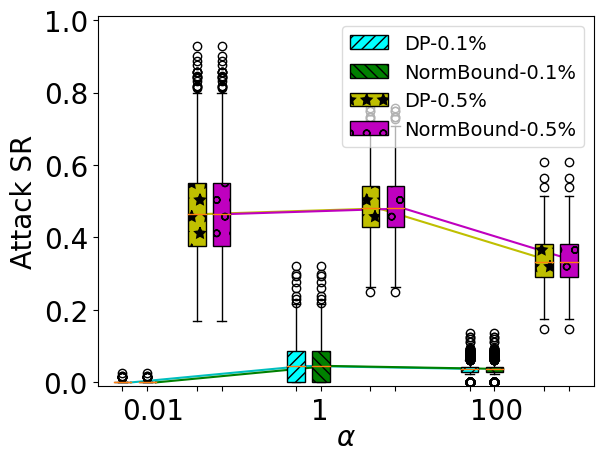}} 
     \subfigure[FedDC-BenignAC]{    \includegraphics[scale=0.27]{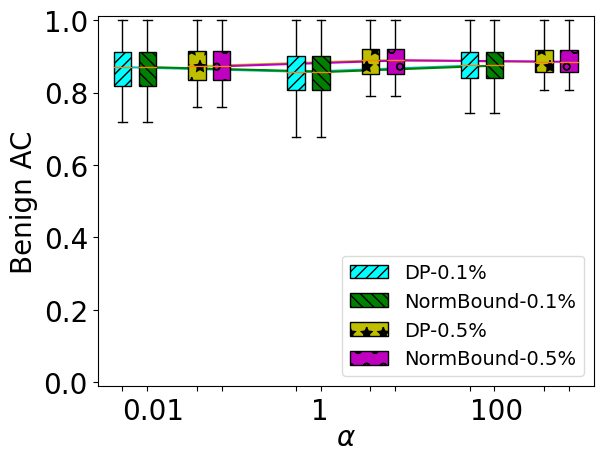} }\hfill
      \subfigure[FedDC-AttackSR]{ \includegraphics[scale=0.27]{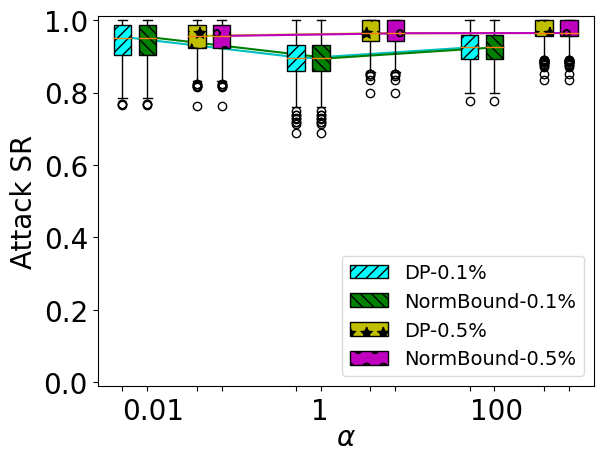}} 
       \subfigure[MetaFed-BenignAC]{ \includegraphics[scale=0.27]{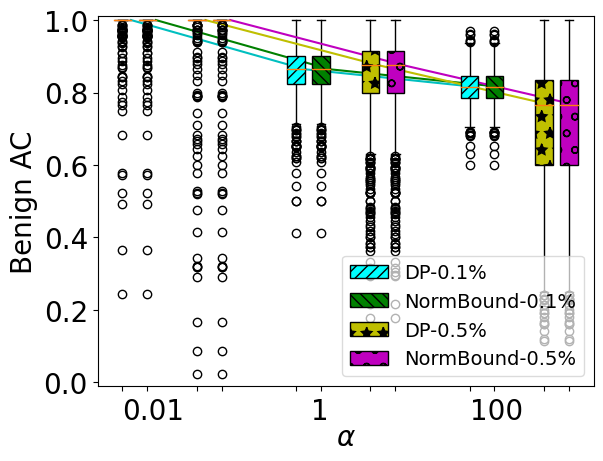} }\hfill
     \subfigure[MetaFed-AttackSR]{  \includegraphics[scale=0.27]{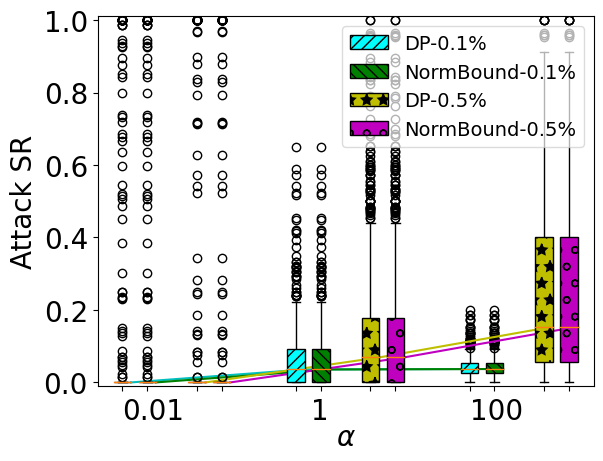}} 
    \caption{ \textsc{CollaPois} (under defenses) with 0.1\% and 0.5\%  compromised clients  for the FEMNIST dataset in FedAvg, FedDC, and MetaFed (Top 25\% Clients). (In MetaFed, many clients have high Attack SR across values of $\alpha$ and defenses. Top-1\% infected clients have an average of Attack SR over 99.5\% (Fig.~\ref{fig:all defenses-femnist-app-0.5-top1},Appx. \ref{app:Supplemental Results}) )} 
    \label{fig:all defenses-femnist-0.1-0.5-mainbody}  
\end{figure}

\end{document}